%% file: ms.tex
\documentclass[sigconf]{acmart}

\usepackage{booktabs} % For formal tables

\copyrightyear{2018}
\acmYear{2018}
\setcopyright{acmcopyright}
\acmConference[MM '18]{2018 ACM Multimedia Conference}{October 22--26, 2018}{Seoul, Republic of Korea}
\acmBooktitle{MM '18: 2018 ACM Multimedia Conference, Oct. 22--26, 2018, Seoul, Republic of Korea}
\acmPrice{15.00}
\acmDOI{10.1145/3240508.3240524}
\acmISBN{978-1-4503-5665-7/18/10}

\usepackage{amsmath}
\usepackage{amssymb}

\usepackage{mathrsfs}
\usepackage{array}

\begin{document}
\settopmatter{printacmref=false, printfolios=false}
\title{Style Separation and Synthesis \\ via Generative Adversarial Networks}
%\titlenote{\footnotesize Corresponding author: Sheng Tang. This work was supported by the National Key Research and Development Program of China (2017YFC0820605), and the National Natural Science Foundation of China (61525206, 61572472, U1703261, 61571424).}
%\subtitle{Extended Abstract}
%\subtitlenote{The full version of the author's guide is available as
%  \texttt{acmart.pdf} document}

\author{
Rui Zhang$^{1,3}$,
Sheng Tang$^{1}$,
Yu Li$^{1,3}$,
Junbo Guo$^{1}$,
Yongdong Zhang$^{1}$,
Jintao Li$^{1}$,
Shuicheng Yan$^{2,4}$ }
\affiliation{
$^1$ Institute of Computing Technology, Chinese Academy of Sciences, Beijing, China. \\
 $^2$  Qihoo 360 Artificial Intelligence Institute, Beijing, China. \\
$^3$ University of Chinese Academy of Sciences, Beijing, China. \\
$^4$ Department of Electrical and Computer Engineering, National University of Singapore, Singapore.  }
\email{{zhangrui, ts, liyu, guojunbo, zhyd, jtli}@ict.ac.cn, yanshuicheng@360.cn}

\begin{abstract}
Style synthesis attracts great interests recently, while few works focus on its dual problem ``style separation''.
In this paper, we propose the \textbf{S}tyle \textbf{S}eparation and \textbf{S}ynthesis \textbf{G}enerative \textbf{A}dversarial \textbf{N}etwork (S$^3$-GAN) to simultaneously implement style separation and style synthesis on object photographs of specific categories. Based on the assumption that the object photographs lie on a manifold, and the contents and styles are independent, we employ S$^3$-GAN to build mappings between the manifold and a latent vector space for separating and synthesizing the contents and styles. The S$^3$-GAN consists of an encoder network, a generator network, and an adversarial network. The encoder network performs style separation by mapping an object photograph to a latent vector. Two halves of the latent vector represent the content and style, respectively. The generator network performs style synthesis by taking a concatenated vector as input. The concatenated vector contains the style half vector of the style target image and the content half vector of the content target image. Once obtaining the images from the generator network, an adversarial network is imposed to generate more photo-realistic images. Experiments on CelebA and UT Zappos 50K datasets demonstrate that the S$^3$-GAN has the capacity of style separation and synthesis simultaneously, and could capture various styles in a single model.
\end{abstract}

%\begin{CCSXML}
%<ccs2012>
%  <concept_id>10010147.10010257.10010293.10010294</concept_id>
%  <concept_desc>Computing methodologies~Neural networks</concept_desc>
%  <concept_significance>500</concept_significance>
%</ccs2012>
%\end{CCSXML}
%
%\ccsdesc[500]{Computing methodologies~Neural networks}

\keywords{Style synthesis; Style separation; Generative Adversarial Network}

\maketitle

%\textbf{ACM Reference Format:}\\
%{\small Rui Zhang, Sheng Tang, Yu Li, Junbo Guo, Yongdong Zhang, Jintao Li, Shuicheng Yan. 2018. Style Separation and Synthesis via Generative Adversarial Networks. In MM'18: 2018 ACM Multimedia Conference, Oct. 22-26, 2018, Seoul, Republic of Korea. ACM, New York, NY, USA, 9 pages. https://doi.org/10.1145/3240508.3240524}

\input{samplebody-conf}

\newpage

\bibliographystyle{ACM-Reference-Format}
\bibliography{sample-bibliography}

\end{document}

%% file: samplebody-conf.tex
\section{Introduction}
Style synthesis~\cite{DBLP:conf/cvpr/GatysEB16}, also known as style transfer and texture synthesis, attracts enormously attentions recently. The goal of style synthesis is to generate a new image which migrates the style (\emph{e.g.} colors, textures) from the style target image, and maintain the content (\emph{e.g.} edges, shapes) of the content target image. Approaches~\cite{DBLP:conf/cvpr/GatysEB16, DBLP:conf/eccv/JohnsonAF16} based on Convolutional Neural Networks (CNNs) \cite{DBLP:conf/nips/KrizhevskySH12, DBLP:journals/corr/SimonyanZ14a} achieve remarkable success on style synthesis and generate astonishing results. Most of those works focus on migrating styles of artistic works to photographs. However, all the objects in photographs have their individual styles, which could also be migrated to other photographs. Moreover, the success of style synthesis shows that the content and style of an image are independent.

Thus how to learn the individual representations for content and style from a given image is the dual problem of style synthesis. We named this problem as ``style separation''. Nowadays, existing works focus on style synthesis and pay a little attention to style separation. For example, methods \cite{DBLP:conf/eccv/JohnsonAF16, DBLP:conf/icml/UlyanovLVL16} could represent styles with the learned feedforward networks, but they cannot represent image contents at the same time.

\begin{figure}[t]
  \centering
  \includegraphics[width=0.95\linewidth]{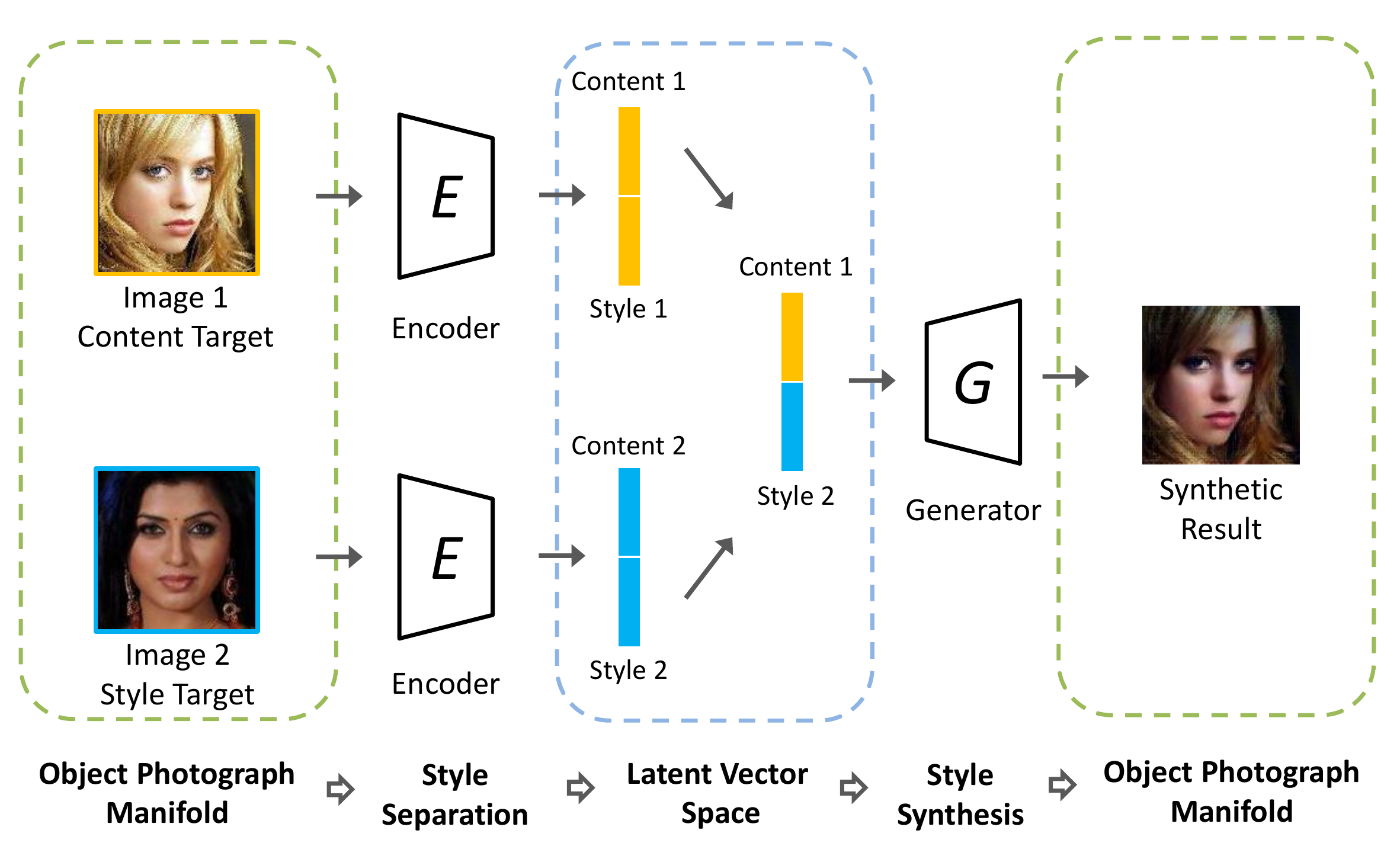}
   \vspace{-10pt}
  \caption{\small The proposed S$^3$-GAN employs a pair of encoder and generator to build mappings between the object photograph manifold and a latent vector space. The encoder performs style separation by encoding the object photograph to the latent vector, half of which represents style, the other half represents content. The generator produces the result of style synthesis from the concatenated vector.}
  \label{fig:motivation}
   \vspace{-10pt}
\end{figure}

In this work, we aim to implement the style separation and style synthesis simultaneously for object photographs. Therefore, we propose a novel network named \textbf{S}tyle \textbf{S}eparation and \textbf{S}ynthesis \textbf{G}enerative \textbf{A}dversarial \textbf{N}etwork (S$^3$-GAN). The S$^3$-GAN is trained on specific categories of objects (\emph{e.g.} faces, shoes, etc.) due to GANs could generate realistic images in specific domains. Inspired by~\cite{DBLP:conf/cvpr/GatysEB16}, we define the structures of objects as ``content'' (\emph{e.g.} the identities and poses of faces, the shapes of shoes), and the colors and textures of objects as ``style'' (\emph{e.g.} the skin color and hair color of faces, the colors and patterns of shoes).
Based on the assumption that the object photographs lie on a high dimensional manifold, S$^3$-GAN employs a pair of encoder and generator to build mappings between the manifold and a latent vector space, as illustrated in Figure \ref{fig:motivation}.
The encoder is used for style separation. At the encoder stage, we map a given photograph to the latent space. As the content and style are independent, we enforce half of the latent vector represents the style, and the other half represents the content.
The generator network performs style synthesis by taking the concatenated vector as input. The concatenated vector contains the style half vector of the style target image and the content half vector of the content target image. The object photograph generated from the concatenated vector has the similar style to the style target image while preserving the content of the content target image.

The proposed S$^3$-GAN shows major differences with existing style synthesis approaches~\cite{DBLP:conf/cvpr/GatysEB16, DBLP:conf/nips/GatysEB15, DBLP:conf/cvpr/LiW16,DBLP:conf/eccv/JohnsonAF16, DBLP:conf/icml/UlyanovLVL16}. Some of them are an iterative optimization method \cite{DBLP:conf/cvpr/GatysEB16, DBLP:conf/nips/GatysEB15, DBLP:conf/cvpr/LiW16}, which can generate high-quality images with high computationally cost. The other approaches employ feedforward networks to generate images closing to the given style target images \cite{DBLP:conf/eccv/JohnsonAF16, DBLP:conf/icml/UlyanovLVL16}. These methods could produce results in real-time, but are only able to handle a specific style one at a time. By comparison, the proposed S$^3$-GAN could handle various styles of objects by a single model, as well as high-efficiently synthesize different styles by concatenating half vectors of different styles and processing forward propagation.

The proposed S$^3$-GAN is derived from GANs, but it has some differences with GANs. The GANs-based methods achieve impressive success in image generation and editing \cite{DBLP:journals/corr/IsolaZZE16, DBLP:conf/icml/KimCKLK17, DBLP:journals/corr/ZhouXYFHH17}. However, most of them build mappings between two application-specific domains for image-to-image translation, which could be regarded as the conversion between two styles. Differently, the proposed S$^3$-GAN could build transfers among various styles. The half vectors of styles could be treated as the conditions to generate images of associated styles. Thus the translations between any two different styles can be simply accomplished by replacing the style half vectors.

We perform the proposed S$^3$-GAN on photographs of two specific categories of objects, faces from CelebA dataset \cite{DBLP:conf/iccv/LiuLWT15} and shoes from UT Zappos50K dataset \cite{DBLP:conf/cvpr/YuG14}. Experimental results show the effectiveness of style separation and synthesis for our proposed method. The main contributions of our work could be summarized as follows:
\begin{itemize}
\item We propose a novel S$^3$-GAN framework for style separation and synthesis. Extensive experiments on photographs of faces and shoes demonstrate the effectiveness of the S$^3$-GAN.
\item The S$^3$-GAN performs style separation with an encoder, which builds the mapping between the object photograph manifold and a latent vector space. For a given object photograph, half of its latent vector is the style representation, and the other half is the content representation.
\item The S$^3$-GAN performs style synthesis with a generator. By concatenating the style half vector of the style target image and the content half vector of the content target image, the generator maps the concatenated vector back to the object photograph manifold to produce the style synthesis result.
\end{itemize}

% ----------------------------------- Related Work --------------------------------------------

\section{Related Work}

\subsection{Style Synthesis}

Style synthesis can be regarded as a generalization of texture synthesis. The previous texture synthesis methods mainly apply low-level image features to grow textures and preserve image structures \cite{DBLP:conf/siggraph/EfrosF01, DBLP:conf/siggraph/HertzmannJOCS01, DBLP:conf/iccv/EfrosL99}.

Recently, approaches based on CNNs generate astonishing results. These approaches employ the perceptual losses measured from CNN features to estimate the style and content similarity of generated images and target images. \cite{DBLP:conf/cvpr/GatysEB16, DBLP:conf/nips/GatysEB15} propose the optimization-based methods to minimize the perceptual losses through an iterative process directly. \cite{DBLP:conf/cvpr/LiW16} extends these works through matching neural patches with Markov Random Fields (MRFs). The optimization-based methods are computationally expensive since the pixel values of synthesis results are gradually optimized from hundreds of backward propagations.

To speed up the process of style synthesis, approaches based on feedforward networks are proposed \cite{DBLP:conf/eccv/JohnsonAF16, DBLP:conf/icml/UlyanovLVL16, DBLP:conf/eccv/LiW16}. These approaches learn feedforward networks to minimize perceptual losses of a specific style target image and any content target images. Therefore, the stylized results of the given photographs can be gained through the forward propagation process, saving the computational time of iterations. However, one model of these methods is only able to represent a single style. For a new style, the feedforward networks have to be retrained.

Until very recently, some approaches attempt to capture multiple styles in a single feedforward network, which represents styles with multiple filter banks \cite{DBLP:journals/corr/ChenYLYH17}, conditional instance normalization \cite{DBLP:journals/corr/DumoulinSK16} or binary selection units \cite{DBLP:conf/cvpr/LiFYWLY17}. There are also some approaches try to represent arbitrary styles in a single model through learning general mappings \cite{Ghiasi2017Exploring}, adaptive instance normalization \cite{DBLP:conf/iccv/HuangB17} or feature transforms \cite{DBLP:conf/nips/LiFYWLY17}.

In this paper, we propose the S$^3$-GAN to implement both style separation and style synthesis. Contents and styles of object photographs are represented as latent vectors. The S$^3$-GAN could not only perform style synthesis through a forward propagation process, but also capture various styles in a single model.

\subsection{Generative Adversarial Networks}

GAN is one of the most successful generative models to generate photorealistic images. The standard GANs \cite{DBLP:conf/nips/GoodfellowPMXWOCB14, DBLP:journals/corr/RadfordMC15} learn a generator and a discriminator from the min-max two-player game. The generator produces plausible images from random noises, while the discriminator distinguishes the generated images from the real samples. The training processes of original GANs are unstable, thus many approaches are proposed for improvement, such as WGAN \cite{DBLP:journals/corr/ArjovskyCB17}, WGAN-GP \cite{DBLP:journals/corr/GulrajaniAADC17}, EBGAN \cite{DBLP:journals/corr/ZhaoML16}, LS-GAN \cite{DBLP:journals/corr/Qi17}.

\begin{figure*}[t]
  \centering
  \includegraphics[width=0.77\linewidth]{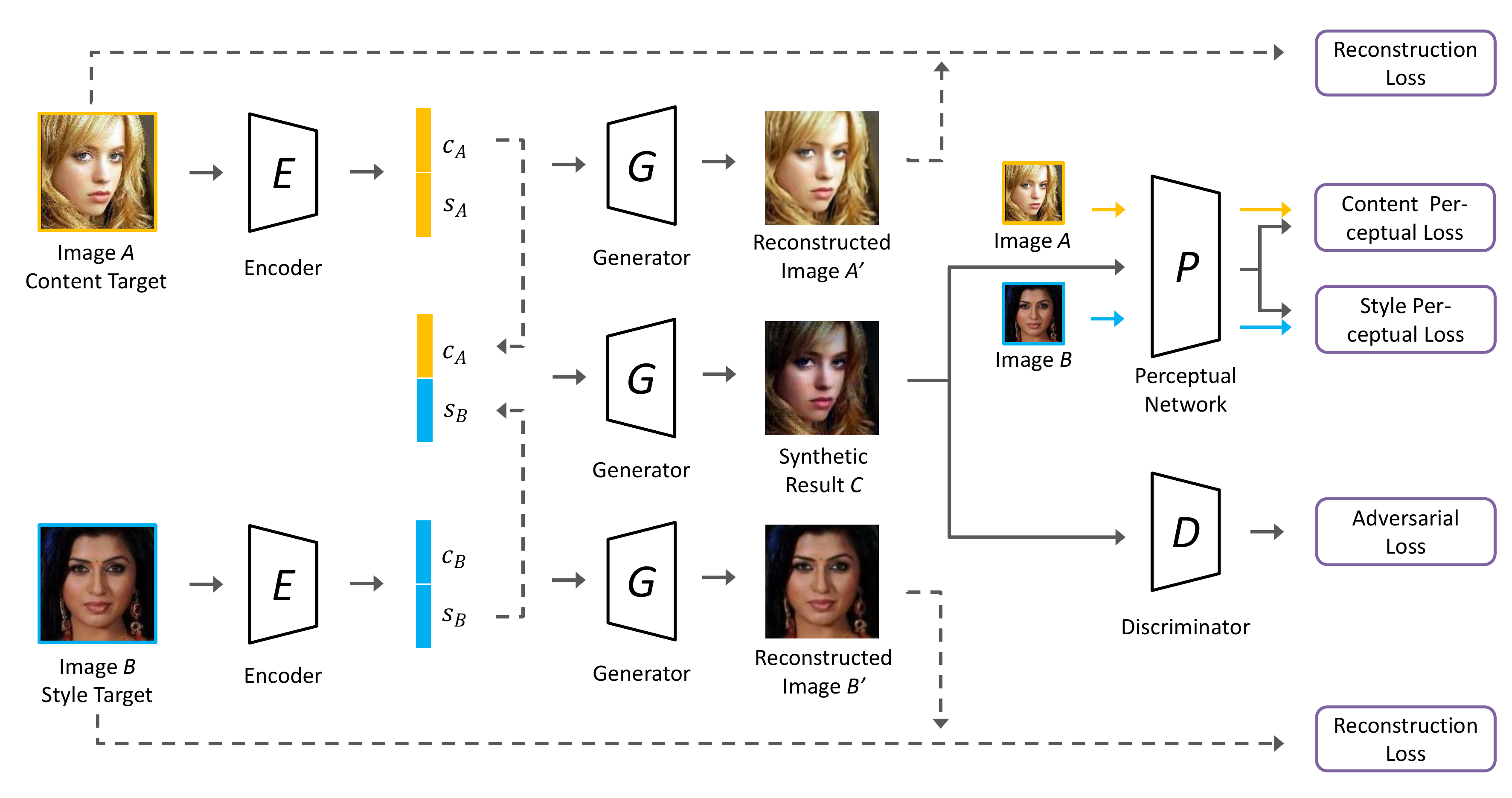}
   \vspace{-10pt}
  \caption{\small The architecture of the proposed S$^3$-GAN consists of the encoder, generator, discriminator and perceptual network.
  The encoder $E$ acquires the representation of style separation by mapping the target images $A$ and $B$ to latent vector $[c_A,s_A]$ and $[c_B,s_B]$. The generator $G$ produces the result $C$ of style synthesis from the concatenated vector $[c_A,s_B]$. The discriminator $D$ evaluates the adversarial loss to help to generate plausible images. The perceptual network $P$ is applied to gain perceptual losses, including content perceptual loss and style perceptual loss. Reconstruction loss and total variation loss are added to the objective function for supplementation (total variation loss is omitted in the figure for simplification).}
  \label{fig:architecture}
   \vspace{-10pt}
\end{figure*}

Moreover, approaches based on Conditional GANs (CGANs) \cite{DBLP:journals/corr/MirzaO14} have been successfully applied to many tasks. These approaches condition GANs on discrete labels \cite{DBLP:journals/corr/MirzaO14}, text \cite{DBLP:conf/icml/ReedAYLSL16} and images. Among them, CGANs conditioned on images accomplish image-to-image translation \cite{DBLP:journals/corr/IsolaZZE16} with an additional encoder, which is introduced to obtain conditions from the input images. These frameworks are widely used to tackle many challenge tasks, such as image inpainting \cite{DBLP:conf/cvpr/PathakKDDE16, DBLP:journals/corr/YangLLSWL16}, super-resolution \cite{DBLP:journals/corr/LedigTHCATTWS16}, age progression and regression \cite{DBLP:journals/corr/ZhangSQ17}, style transfer \cite{DBLP:journals/corr/ZhuPIE17}, scene synthetic \cite{DBLP:conf/eccv/WangG16}, cross-modal retrieval \cite{DBLP:conf/ijcai/ChiP18, DBLP:conf/aaai/ZhangPY18, DBLP:conf/mm/YaoZZLT17} and face attribute manipulation \cite{DBLP:journals/corr/ShenL16b, DBLP:conf/icml/KimCKLK17, DBLP:journals/corr/ZhouXYFHH17}. Moreover, approaches of domain-adaptation \cite{DBLP:journals/corr/abs-1802-10349, DBLP:journals/corr/abs-1804-08286} employ GANs to adapt features and boost models of traditional tasks, such as semantic segmentation. Some other approaches \cite{DBLP:conf/nips/MaJSSTG17, DBLP:journals/corr/abs-1801-00055} utilize GANs to generate human images of arbitrary poses and benefit the related tasks such as person re-identification. Most of these approaches perform image-to-image translation by building mappings between two application-specific domains.

In this paper, the proposed S$^3$-GAN are trained to represent the domain consisting of object photographs. These domains can be divided into many sub-domains by different styles. The S$^3$-GAN could perform mappings between any pair of sub-domains to accomplish arbitrary style transfer.

% -------------------------------------- Methods --------------------------------------------

\section{Proposed Approaches}

In this section, we first formulate the latent vector space which is introduced to disentangle the content and style representations. Then we demonstrate the pipeline of S$^3$-GAN and describe each component in detail. Finally, we present all the individual loss functions utilized to optimize the S$^3$-GAN.

\subsection{Formulation}

We assume the object photographs of a specific category lie on a high dimensional manifold $\mathcal{M}$ in the photograph domain. Objects with same styles or same contents will be clustered to the sub-domains of associated styles or contents.

Since it is difficult to directly model photographs in the manifold $\mathcal{M}$, we build mappings between manifold $\mathcal{M}$ and a latent vector space $\mathcal{L}\subset\mathbb{R}^{2d\times k\times k}$, where $d,k\in\mathbb{Z}^{+}$ represent the dimensions of vectors in $\mathcal{L}$. Considering contents and styles are independent, we attempt to disentangle the representations of contents and styles to different dimensionality of the latent vectors in $\mathcal{L}$. Suppose for a given object photograph $I\in\mathcal{M}$, its associated latent vector in $\mathcal{L}$ is $[c_I, s_I]$, where $c_I\in\mathbb{R}^{d\times k\times k}$ and $s_I\in\mathbb{R}^{d\times k\times k}$ are the sub-vectors representing its content and style respectively. We set the sub-vectors of content and style with equal dimensionality for simplification. Therefore, $[\cdot, s_I]$ (or $[c_I, \cdot]$) could represent the sub-domain containing all the objects showing different contents (or styles) but the same style (or content) with $I$. For any style sub-vector $\hat{s}$, $[c_I, \hat{s}]$ is the intersection of the sub-domain of style $\hat{s}$ and the sub-domain $[c_I, \cdot]$. Thus, $[c_I, \hat{s}]$ could represent the result of modifying style to $\hat{s}$ while preserving the content of $I$.

\subsection{Architecture}

The proposed S$^3$-GAN employs the framework of GANs to learn the mapping from the manifold $\mathcal{M}$ to the latent vector space $\mathcal{L}$, as well as generate realistic images from $\mathcal{L}$. The pipeline of the S$^3$-GAN consists of four components, including the encoder, generator, discriminator and perceptual network, as shown in Figure \ref{fig:architecture}. The encoder and generator are applied both in the training and test stages to perform style separation and style synthesis, while the discriminator and perceptual network are employed only in the training stage to optimize the objective functions.

We learn the encoder $ E: \mathcal{M} \to \mathcal{L}$ to build the mapping from the manifold $\mathcal{M}$ to the latent vector space $\mathcal{L}$. For any content target image $A$ and style target image $B$, their corresponding latent vectors in $\mathcal{L}$ are denoted as:
\begin{equation}\label{equ1}
[c_A,s_A]=E(A), A\in\mathcal{M}, [c_A,s_A]\in\mathcal{L},
\end{equation}
\begin{equation}\label{equ2}
[c_B,s_B]=E(B), B\in\mathcal{M}, [c_B,s_B]\in\mathcal{L}.
\end{equation}
Thus, style separation could be implemented by $E$. $c_A$ (or $c_B$) is the representation of content and $s_A$ (or $s_B$) is the representation of style for the object photograph $A$ (or $B$).

On the contrary, we also learn the generator $ G: \mathcal{L} \to \mathcal{M}$ to build the mapping from latent vector space $\mathcal{L}$ back to manifold $\mathcal{M}$:
\begin{equation}\label{equ3}
A'=G([c_A,s_A]), A,A'\in\mathcal{M}, [c_A,s_A]\in\mathcal{L},
\end{equation}
\begin{equation}\label{equ4}
B'=G([c_B,s_B]), B,B'\in\mathcal{M}, [c_B,s_B]\in\mathcal{L},
\end{equation}
where $A'$ (or $B'$) is the reconstruction having the same content and style of $A$ (or $B$). The latent sub-vectors of content and style target images could be utilized as the conditions to generate results of style synthesis. Therefore, the generator $G$ could produce a synthetic photograph $C$ from concatenating the associated sub-vectors $c_A$ from the content target $A$ and $s_B$ from the style target $B$:
\begin{equation}\label{equ5}
C=G([c_A,s_B]), C\in\mathcal{M}, [c_A,s_B]\in\mathcal{L}.
\end{equation}

Inspired by the framework of GANs, we also introduce the discriminator $D$ to classify whether an image is real or fake (\emph{i.e.,} produced by the generator). The synthetic result $C$ and real photographs randomly sampled from the training set are fed into the discriminator $D$ to acquire the adversarial loss. Indistinguishable object photographs will be generated during the optimal process of the min-max game.

Moreover, we bring the perceptual network $P$ to evaluate and improve style synthesis results. $P$ is employed to extract features from the synthetic result $C$, the content target $A$ and the style target $B$ to evaluate the perceptual losses, including content perceptual loss and style perceptual loss. The perceptual losses enforce $C$ to acquire the style of $B$ while preserving the content of $A$.

\subsection{Loss Functions}

Figure \ref{fig:architecture} also presents the losses for optimizing the proposed S$^3$-GAN. The objective function is the weighted sum of five losses, including adversarial loss, content perceptual loss, style perceptual loss, reconstruction loss and total variation loss. They will be described in detail in the following.

\subsubsection{Adversarial Loss}

We apply the discriminator $D$ to evaluate the adversarial loss $L_{A}$.
The adversarial loss of original GAN \cite{DBLP:conf/nips/GoodfellowPMXWOCB14} is based on the Kullback-Leibler (KL) divergence. However, when the discriminator is quickly trained towards its optimality, the KL divergence will lead to a constant and cause the vanishing gradient problem, which will restrain the updating of the generator. To tackle this problem, we exploit the adversarial loss with the recently proposed WGAN \cite{DBLP:journals/corr/ArjovskyCB17}, which is based on the Earth Mover (EM) distance.
%We do not utilize the further improved WGAN-GP \cite{DBLP:journals/corr/GulrajaniAADC17}, since we already gain satisfactory results from WGAN and it is computationally expensive to calculate the gradient penalty in WGAN-GP.

We denote the distribution of the training data (\emph{i.e.,} the object photographs of specific categories) in the manifold $\mathcal{M}$ as $p_{\mathcal{M}}$. Random sampling process from $p_{\mathcal{M}}$ is denoted as $I\sim p_{\mathcal{M}}$. Thus, the adversarial loss is:
\begin{equation}\label{equ8}
\begin{array}{l}
L_{A}(E,G,D) =  \mathbb{E}_{I \sim p_{\mathcal{M}}}\left[D(I)\right] \\
 \qquad\qquad -  \mathbb{E}_{A,B \sim p_{\mathcal{M}}}\left[D(G([c_A,s_B]))\right], \\
\end{array}
\end{equation}
where $G([c_A,s_B])$ is the generated result of style synthesis, formulated in Eq.~\eqref{equ1}, Eq.~\eqref{equ2} and Eq.\eqref{equ5}.
A min-max objective function is employed to optimize the adversarial loss:
\begin{equation}\label{equ7}
arg \min_{E,G} \max_{D} L_{A}(E,G,D),
\end{equation}
where $E,G$ tries to minimize $L_{A}$ so as to generate image $G([c_A,s_B])$ that looks indistinguishable to images from training set, while $D$ tries to maximize $L_{A}$ so as to classify the generated image $ G([c_A,s_B])$ and real sample $I$.

The adversarial loss ensures that the generated images reside in the manifold $\mathcal{M}$, and forces them to be indistinguishable from real images.
Thus, we exploit this loss function to produce realistic images. Blurry images look obviously fake, so that they will be prevented by the adversarial loss.

\subsubsection{Content Perceptual Loss}

The generated image $C$ are purposed to be stylistically similar to the style target $B$ and preserve the content of the content target $A$. Since the groundtruths of style synthesis are not provided in the training set, we employ the perceptual network $P$ and utilize the feature representations to penalize the differences between generated images and target images, by incorporating the prior knowledge of style synthesis \cite{DBLP:conf/cvpr/GatysEB16, DBLP:conf/eccv/JohnsonAF16}.

Generated results are expected to match the feature responses of the target images.
Let $P_l(I)$ be the feature maps extracted from layer $l$ of the perceptual network $P$ and the input image $I\in\mathcal{M}$. The content perceptual loss $L_C$ is defined as the squared Euclidean distance between feature responses:
\begin{equation}\label{contentloss}
L_C(E,G) = \sum_{l\in\mathcal{F}_C} ||P_l(C)-P_l(A)||_2^2,
\end{equation}
where $\mathcal{F}_C$ are the layers utilized to evaluate the content perceptual loss.

Considering the design of neural networks, higher layers capture semantic-level information including shapes and spatial structures but ignoring low-level information such as colors and textures. Therefore, we calculate $L_C(E,G)$ on higher layers of $P$, so that the generated image $C$ will preserve the content of the content target $A$.

\subsubsection{Style Perceptual Loss}

Suppose the feature map $P_l(I)$ from layer $l$ of input $I$ has the shape of $C_l\times H_l\times W_l$. The style perceptual loss $L_S$ is calculated with the squared Frobenius distance of Gram matrix, denoted as:
\begin{equation}\label{styleloss}
L_S(E,G) = \sum_{l\in\mathcal{F}_S} ||\psi_l(C)-\psi_l(B)||_F^2,
\end{equation}
where $\mathcal{F}_S$ is the layers applied for the style perceptual loss. The Gram matrix $\psi_l(I)$ is a $C_l\times C_l$ matrix inspired from the uncentered covariance of the feature map $P_l(I)$ along the channel dimension. Its element at $(c,c')$ is denoted as:
\begin{equation}\label{}
\psi_l^{c,c'}(I) = \frac{1}{C_lH_lW_l}\sum_{h,w}P_l^{h,w,c}(I)P_l^{h,w,c'}(I).
\end{equation}
The Gram matrix focuses on features activating together from different channels, omitting the spatial information of images. Thus, the style perceptual loss $L_S(E,G)$ based on Gram matrix maintains the style of the style target $B$ and ignores the content. In contrast to the content perceptual loss, $L_S(E,G)$ are calculated on lower layers of $P$ to focus on low-level information including style-related colors and textures.

\subsubsection{Reconstruction Loss}

We could also gain the reconstruction of $A$ and $B$, as formulated in Eq.\eqref{equ3} and Eq.\eqref{equ4}. The reconstruction loss $L_R$ calculated from the original images and the reconstructed images is added to the objective function for supplementation, denoted as:
\begin{equation}\label{reconloss}
\begin{array}{l}
L_{R}(E,G) =||A'-A||_1+||B'-B||_1 \\
=||G(E(A))-A||_1+||G(E(B))-B||_1. \\
\end{array}
\end{equation}
We apply L1 distance rather than L2 in $ L_{R}$, because L1 results in less blurring.

The reconstruction loss ensure that the encoder $E$ and the generator $G$ are a pair of inverse mappings to each other. Considering the groundtruth of style synthesis are not given in the training set, the implementation of reconstruction could also provide an analogous groundtruth output, so as to accelerate the training process and improve the realistic effect. In addition, although the reconstruction loss may overly smooth and lead to blurry images, serious results could be prevented with an appropriate loss weight and the restrict of the adversarial loss.

\subsubsection{Total Variation Loss}

Another auxiliary loss function is the total variation loss $L_{TV}$, which could encourage spatial smoothness of generated results and reduce spike artifacts. It performs total variation regularizer on both the synthesis products and the reconstruction results, formulated as:
\begin{equation}\label{}
\phi(\textbf{x}) = \sum_{i,j}((x_{i,j+1}-x_{i,j})^2+(x_{i+1,j}-x_{i,j})^2),
\end{equation}
\begin{equation}\label{}
\begin{array}{l}
L_{TV}(E,G) = \phi(C) + \phi(A') + \phi(B') \\
 = \phi(G([c_A,s_B])) + \phi(G(E(A))) + \phi(G(E(B))). \\
\end{array}
\end{equation}

\subsubsection{Full Objective Function}

The full objective function $L_O$ is the weighted sum of all the losses defined above, denoted as:
\begin{equation}\label{fullobj}
\begin{array}{l}
L_{O}(E,G,D) = \lambda_1 L_{A}(E,G,D) + \lambda_2 L_{C}(E,G) \\
\qquad + \lambda_3 L_{S}(E,G) + \lambda_4 L_{R}(E,G) + \lambda_5 L_{TV}(E,G),\\
\end{array}
\end{equation}
where $\lambda_1, \lambda_2, \lambda_3, \lambda_4, \lambda_5$ are the loss weights which control the relative importance in the objective function. The optimizing process is to solve the min-max problem:
\begin{equation}\label{}
E^*,G^* = arg\min_{E,G}\max_{D}L_{O}(E,G,D).
\end{equation}

\section{Experiments}

In this section, we perform experiments on object photographs of two specific categories, including faces from CelebA dataset \cite{DBLP:conf/iccv/LiuLWT15} and shoes from UT Zappos50K dataset \cite{DBLP:conf/cvpr/YuG14}.

\subsection{Experimental Settings}

\subsubsection{CelebA Dataset}

The CelebA dataset consists of more than 200K celebrity images of 10K identities. We crop the 128$\times$128 center part of the aligned face images in the CelebA dataset for pre-processing. We randomly select 2K images for testing, and the rest images are employed as training samples. The content and style target image pairs are randomly selected, while the forty face attributes and five key points annotated in the CelebA dataset are not utilized.

\subsubsection{UT Zappos50K Dataset}

The UT Zappos50K dataset is collected from the online shopping website Zappos.com. This dataset contains 50K catalog shoe images which are pictured in the same orientation with blank backgrounds. The images are scaled to 128$\times$128 before being fed to the network. We randomly split these images into two parts, one contains 2K images for testing and the other contains 48K for training. We also randomly select the content and style target image pairs and ignore the meta-data (\emph{e.g.,} shoe type, materials, gender, etc.) of the images.

\begin{table}[t]
\caption{\small The detailed structure of S$^3$-GAN, including encoder $E$, generator $G$ and discriminator $D$. BN: Batch Normalization.}
\label{tab:net}
\vspace{-5pt}
\centering
\begin{tabular}{ccc}
%\hline\hline
\toprule[1pt]
 & \textbf{Encoder} $E$& \\
\hline
Layer						&	Filter Size	| Stride			&	Activation Size	\\
\hline
Input color image			&	-						&	3$\times$128$\times$128	\\
Conv, BN, Leaky ReLU	&	64$\times$4$\times$4$|$2	&	64$\times$64$\times$64	\\
Conv, BN, Leaky ReLU	&	128$\times$4$\times$4$|$2	&	128$\times$32$\times$32	\\
Conv, BN, Leaky ReLU	&	256$\times$4$\times$4$|$2	&	256$\times$16$\times$16	\\
Conv, BN, Leaky ReLU	&	512$\times$4$\times$4$|$2	&	512$\times$8$\times$8	\\
Conv, BN, Leaky ReLU	&	1024$\times$4$\times$4$|$2	&	1024$\times$4$\times$4	\\
%\hline\hline
\midrule[1pt]
 &\textbf{Generator} $G$& \\
\hline
Layer					&	Filter Size				&	Activation Size	\\
\hline
Input latent vector		&	-						&	1024$\times$4$\times$4	\\
Deconv, BN, ReLU	&	512$\times$4$\times$4$|$2	&	512$\times$8$\times$8	\\
Deconv, BN, ReLU	&	256$\times$4$\times$4$|$2	&	256$\times$16$\times$16	\\
Deconv, BN, ReLU	&	128$\times$4$\times$4$|$2	&	128$\times$32$\times$32	\\
Deconv, BN, ReLU	&	64$\times$4$\times$4$|$2	&	64$\times$64$\times$64	\\
Deconv, Tanh			&	3$\times$4$\times$4$|$2	&	3$\times$128$\times$128	\\
%\hline\hline
\midrule[1pt]
 &\textbf{Discriminator} $D$ & \\
\hline
Layer						&	Filter Size				&	Activation Size	\\
\hline
Input color image			&	-						&	3$\times$128$\times$128	\\
Conv, Leaky ReLU			&	64$\times$4$\times$4$|$2	&	64$\times$64$\times$64	\\
Conv, BN, Leaky ReLU	&	128$\times$4$\times$4$|$2	&	128$\times$32$\times$32	\\
Conv, BN, Leaky ReLU	&	256$\times$4$\times$4$|$2	&	256$\times$16$\times$16	\\
Conv, BN, Leaky ReLU	&	512$\times$4$\times$4$|$2	&	512$\times$8$\times$8	\\
Conv, BN, Leaky ReLU	&	1024$\times$4$\times$4$|$2	&	1024$\times$4$\times$4	\\
Fully Connected				&	1$\times$(1024$\times$4$\times$4)	&	1	\\
%\hline\hline
\bottomrule[1pt]
\end{tabular}
   \vspace{-10pt}
\end{table}

\subsubsection{Implementation Details}

The detailed structures of the encoder $E$, generator $G$ and discriminator $D$ are specified in Table \ref{tab:net}. $E$ and $D$ apply the ``Convolution, Batch Normalization, Leaky ReLU'' module, while $G$ exploits the ``Deconvolution, Batch Normalization, ReLU'' module. Strides of 2 are utilized in both convolutional and deconvolutional layers to down-sample or up-sample the feature maps. Particularly, the output layer of $G$ utilizes Tanh as activation function instead of ReLU. In addition, Batch Normalization \cite{DBLP:conf/icml/IoffeS15} is removed in the generator output layer and the discriminator input layer, since directly applying Batch Normalization to all layers may lead to sampling oscillation and model instability. Images is normalized to [0,1] before input to $E$ or $D$, and the output of $G$ are re-scaled to [0,255]. For a given image $I$, the output 1024$\times$4$\times$4 latent vector of $E(I)$ is split along the channel dimension as $c_I$ and $s_I$, \emph{i.e.,} the two half 512$\times$4$\times$4 vectors are the content and style representations respectively.

During training, the Adam optimizer \cite{DBLP:journals/corr/KingmaB14} with the mini-batch of 16 samples is adopted. We employ VGG-19 network \cite{DBLP:journals/corr/SimonyanZ14a} pre-trained on ImageNet dataset \cite{DBLP:conf/cvpr/DengDSLL009} as the perceptual network $P$. The weights of encoder $E$, generator $G$ and discriminator $D$ are initialized from a zero-centered Gaussian distribution with appropriate deviations \cite{DBLP:journals/jmlr/GlorotB10}. The learning rate is fixed at 0.001 for 30 epochs. The loss weights are set as $\lambda_1=1, \lambda_2=10^{-6}, \lambda_3=5\times 10^{-5}, \lambda_4=30, \lambda_5=1$. The absolute values of loss weights are obtained and chosen from the experiments. The direct loss values of perceptual losses are much larger than other losses, so we set the loss weights of perceptual losses smaller than other losses to make the weighted losses in the same order of magnitude. We compute content perceptual loss at layer relu4\_2 and style perceptual loss at layers relu1\_1, relu2\_1 and relu3\_1.
We perform the alternative training approach of GANs, by alternating between one gradient descent step on $D$ and two steps on $E$ and $G$. Our experiments are implemented based on Tensorflow platform. All of our networks are trained and tested on one NVIDIA Tesla K40 GPU.

\begin{figure}[t]
  \centering
  \includegraphics[width=\linewidth]{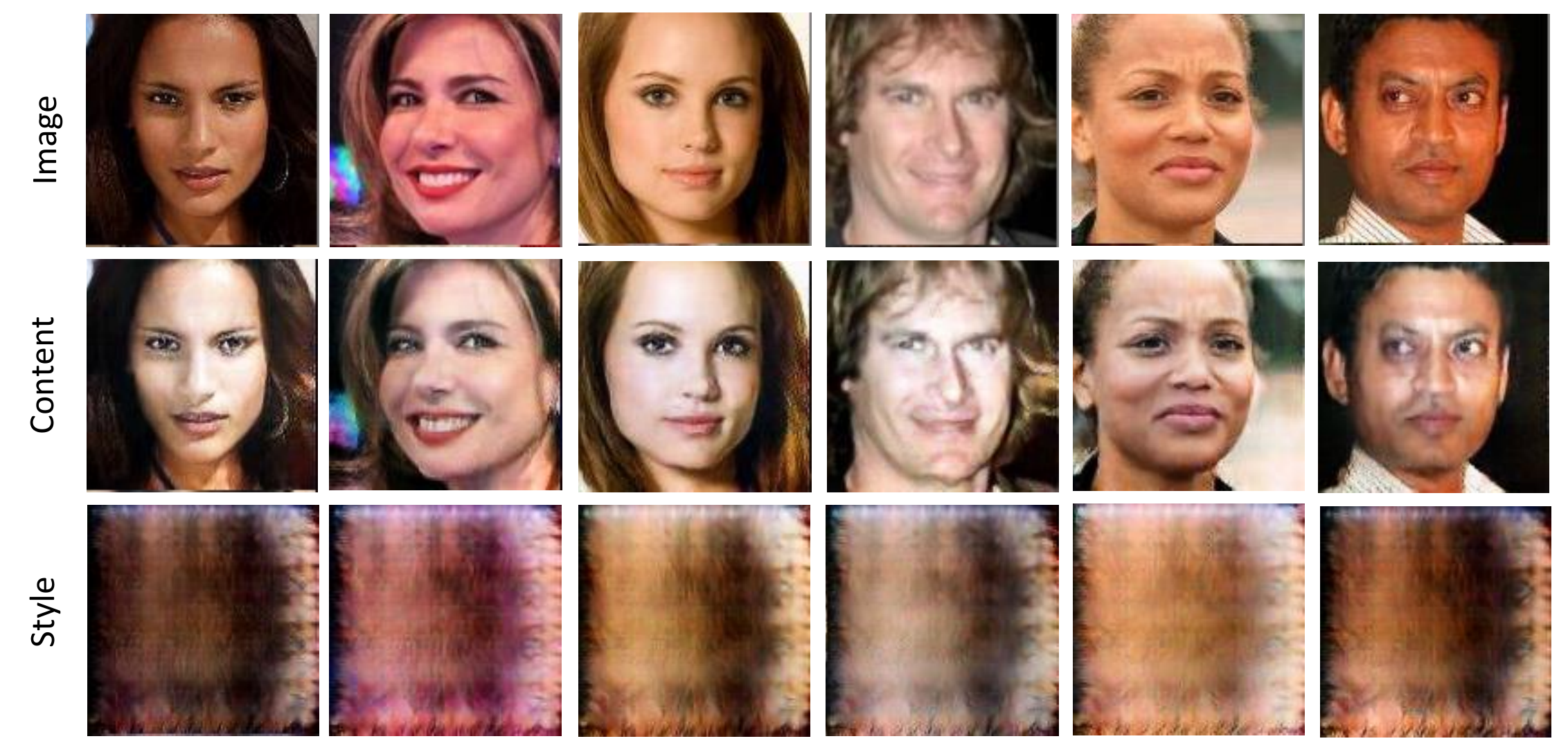}
   \vspace{-12pt}
  \caption{\small Visualization of the content and style representation on face images of CelebA Dataset. From top to bottom are: original images, content representation and style representation.}
  \label{fig:visual_face}
   \vspace{-12pt}
\end{figure}

\begin{figure}[t]
  \centering
  \includegraphics[width=\linewidth]{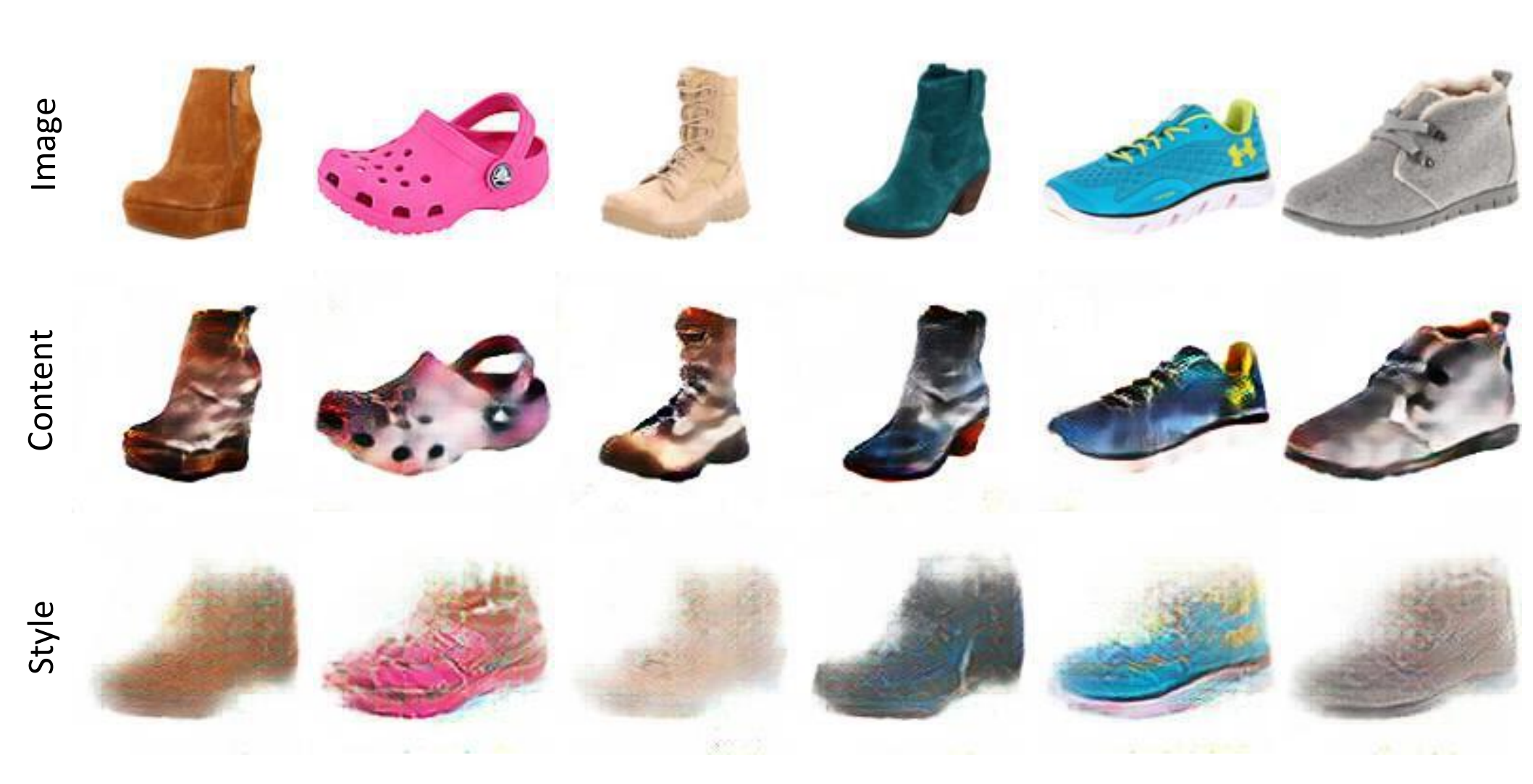}
   \vspace{-20pt}
  \caption{\small Visualization of the content and style representation on shoe images of UT Zappos50K Dataset. From top to bottom are: original images, content representation and style representation.}
  \label{fig:visual_shoe}
   \vspace{-12pt}
\end{figure}

\subsection{Results and Comparisons}

\subsubsection{Visualization of Content and Style Representations}

The encoder of S$^{3}$-GAN could perform style separation through encoding an image to a latent vector, half of which represents style, and the other half represents content. To demonstrate the S$^{3}$-GAN has the capacity of  style separation, we visualize the content and style representations produced by the encoder $E$. For visualization, we preserve the half vector of content or style and simply fill the other half vector with zeros. Then we feed the new vector into the generator $G$, and obtain the visualization for content or style. As shown in Figure \ref{fig:visual_face} and \ref{fig:visual_shoe}, the visualizations of content representations preserve the structure information but abandon color information, while the style representations maintain the color information but ignore structure information. For example, the content representations preserve the identities and poses of original faces in Figure \ref{fig:visual_face}, and the shapes and structures of original shoes in Figure \ref{fig:visual_shoe}.
In contrast, the style representations present the skin color and hair color of style target faces in Figure \ref{fig:visual_face}, and the colors and textures of style target shoes in Figure \ref{fig:visual_shoe}.
These content and style representations are powerful to synthesizing new images.
The above experiments show that the content and style representations are complementary and could be captured from the learned encoder.

\begin{figure}[t]
  \centering
  \includegraphics[width=\linewidth]{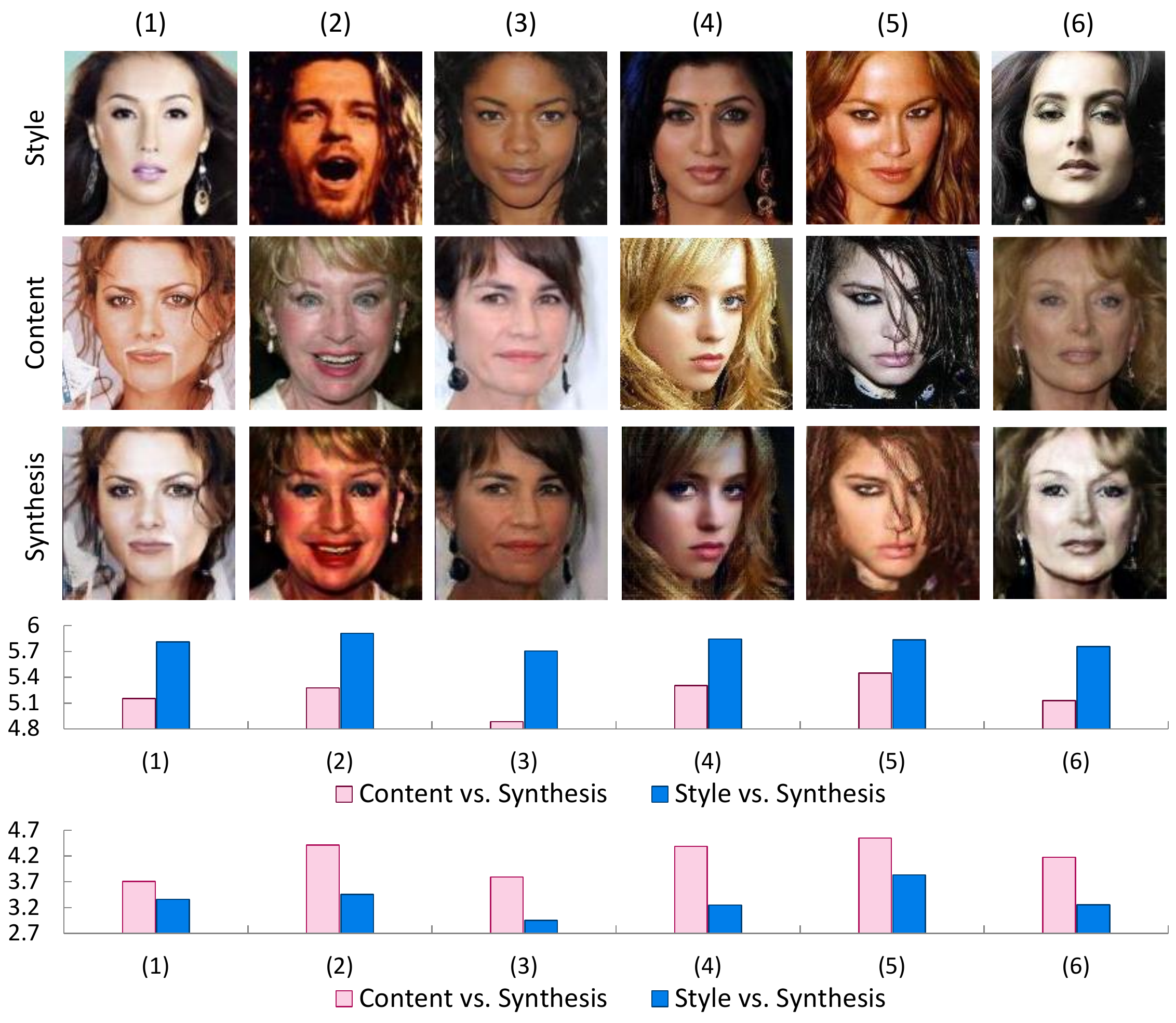}
   \vspace{-15pt}
  \caption{\small Illustration of style synthesis on CelebA dataset. From top to bottom: style target images, content target images, synthesis results, logarithms of content distances and style distances. For the fourth and fifth rows, pink bars are distances between content target images and synthesis results, while blue bars are distances between style target images and synthesis results.}
  \label{fig:pairface}
   \vspace{-15pt}
\end{figure}

\begin{figure}[t]
  \centering
  \includegraphics[width=\linewidth]{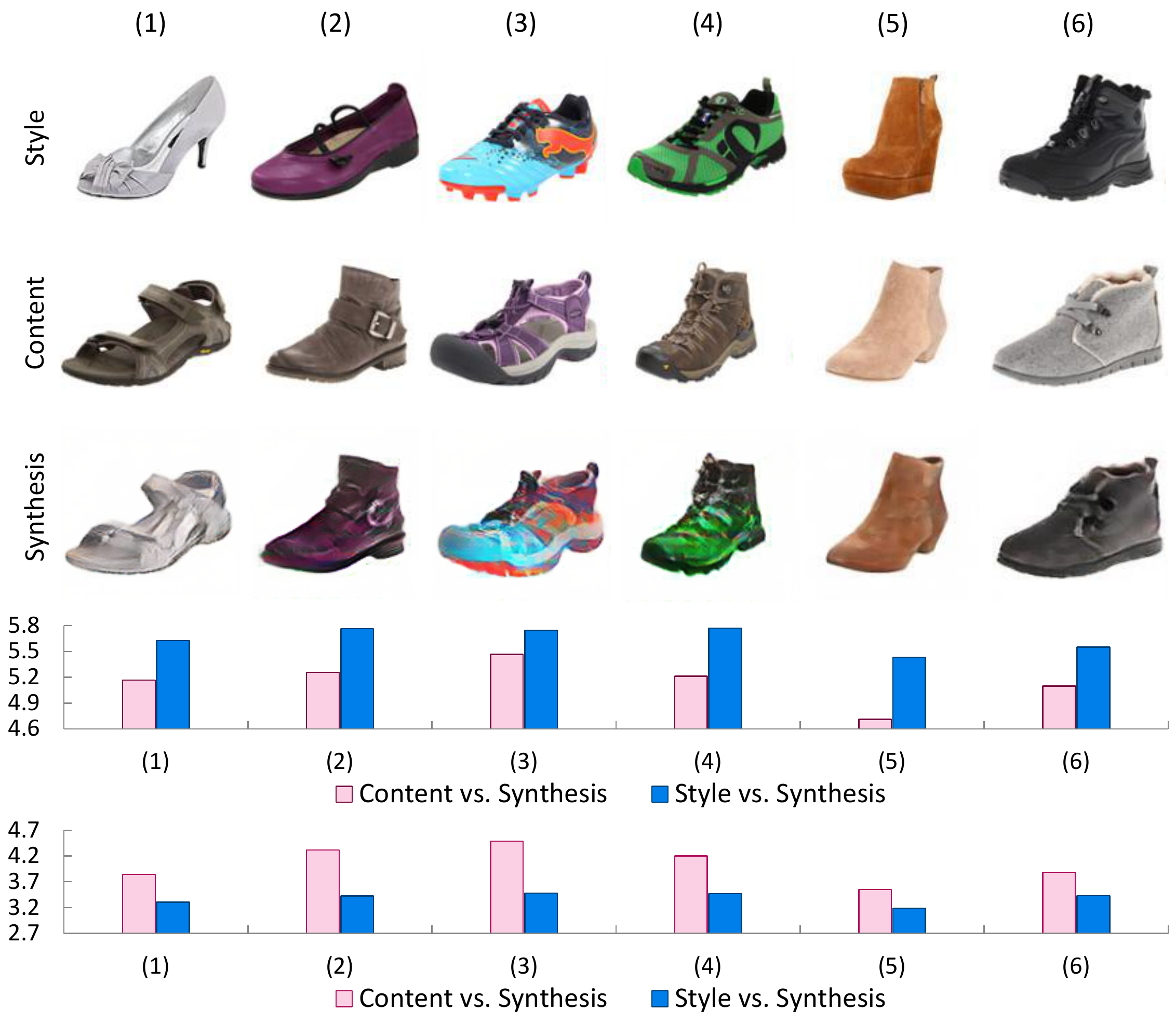}
   \vspace{-15pt}
  \caption{\small Illustration of style synthesis on UT Zappos50K dataset. From top to bottom: style target images, content target images, synthesis results, logarithms of content distances and style distances. For the fourth and fifth rows, pink bars are distances between content target images and synthesis results, while blue bars are distances between style target images and synthesis results.}
  \label{fig:pairshoe}
   \vspace{-12pt}
\end{figure}

 \begin{figure}[t]
   \centering
   \includegraphics[width=0.95\linewidth]{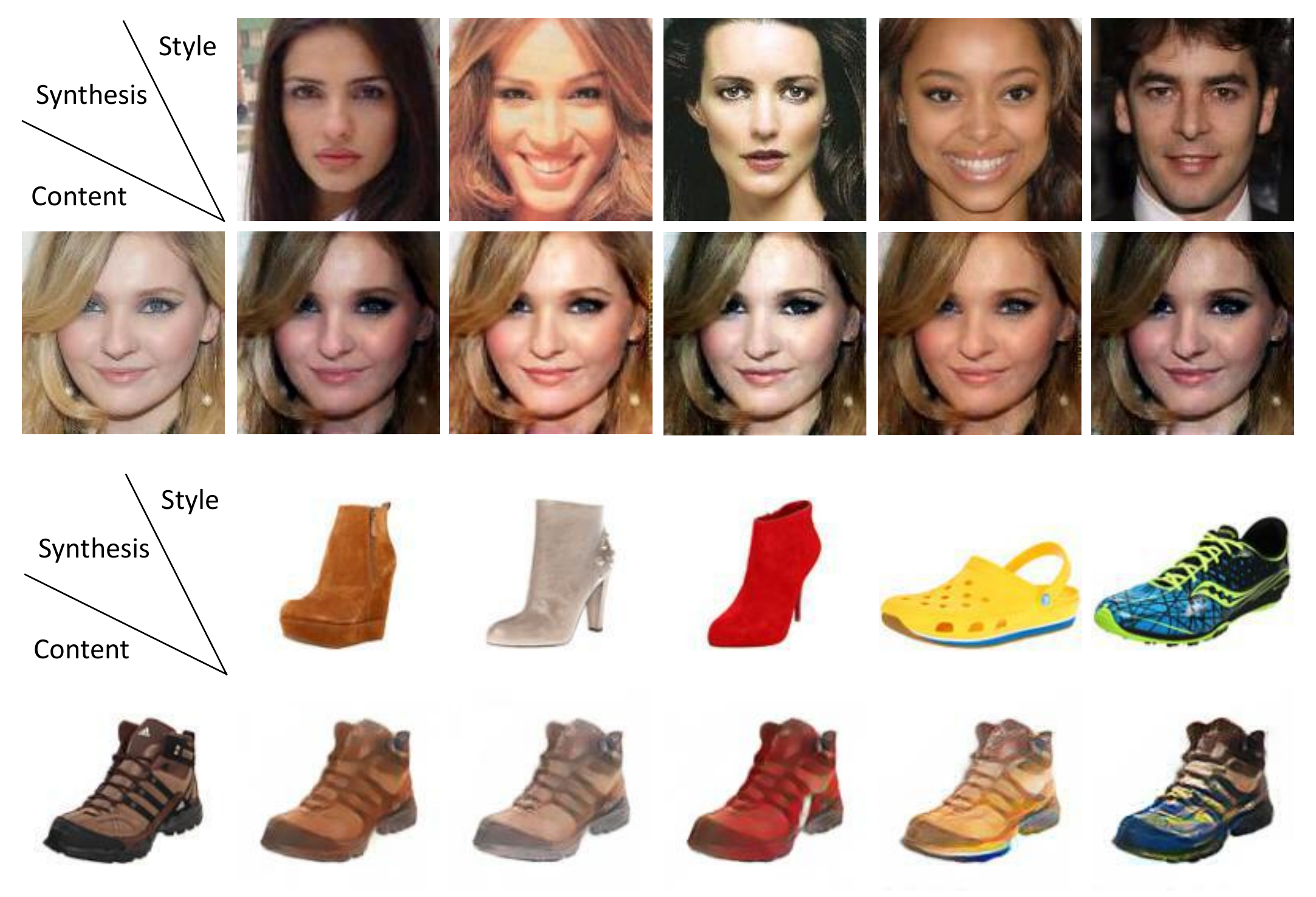}
   \vspace{-10pt}
   \caption{\small Illustration of style synthesis on different styles. The first row shows different style target images, while the second row shows the content target image and style synthesis results. }
   \label{fig:samplecontent}
   \vspace{-15pt}
 \end{figure}

 \begin{figure}[t]
   \centering
   \includegraphics[width=0.95\linewidth]{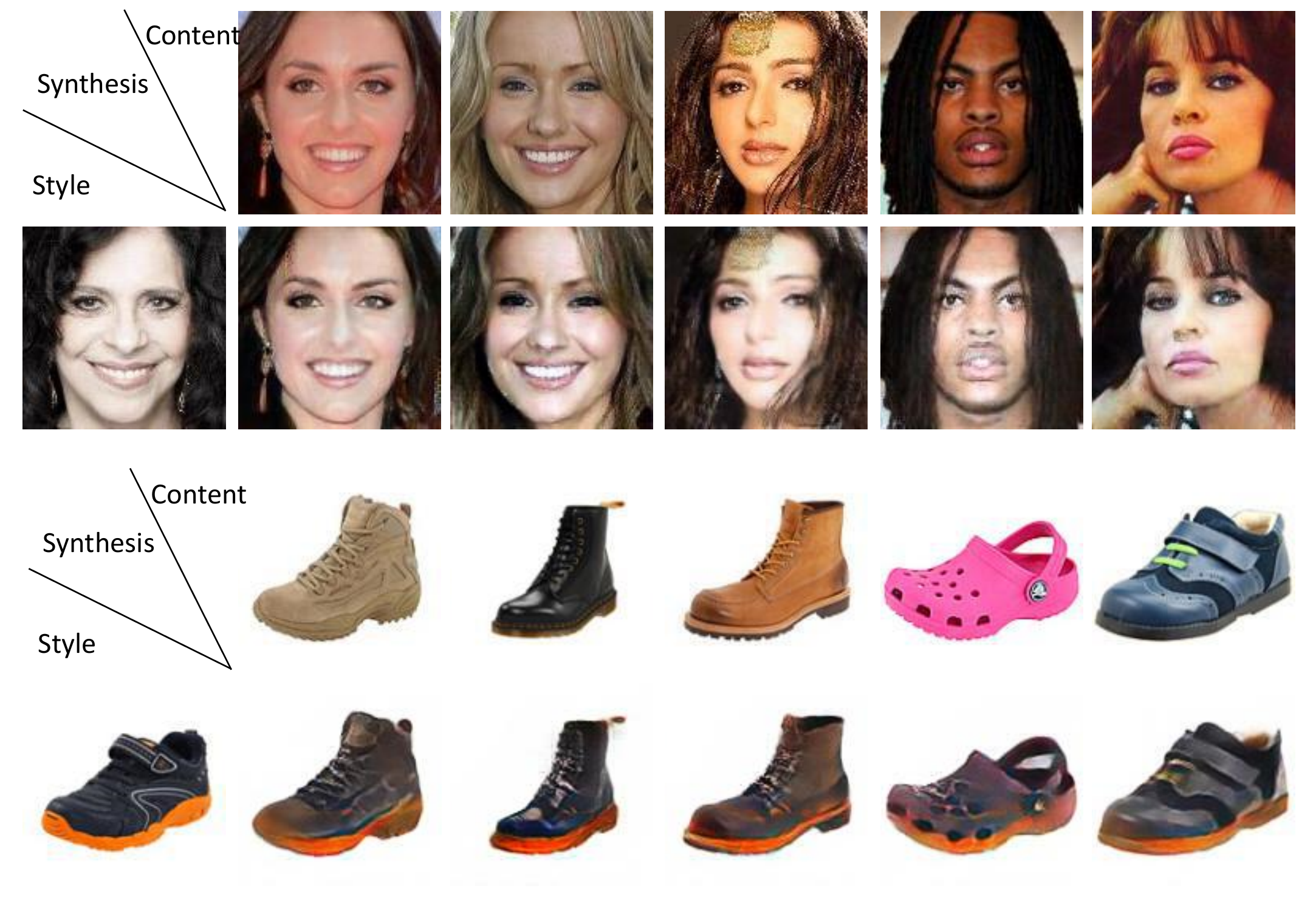}
   \vspace{-10pt}
   \caption{\small Illustration of style synthesis on different contents. The first row shows different content target images, while the second row shows the style target image and style synthesis results. }
   \label{fig:samplestyle}
   \vspace{-10pt}
 \end{figure}

\subsubsection{Results of Style Synthesis}
The generator of S$^3$-GAN could produce style synthesis from the vector by concatenating content and style half vectors.
Both qualitative and quantitative evaluation of synthesis results on CelebA and UT Zappos50K datasets are presented in Figure \ref{fig:pairface} and \ref{fig:pairshoe}.
From the first three rows of those two figures, we could observe that the synthesis images represent the obvious style of the style target images and preserve the content of the content target images. For example, the synthesis faces in the third row in Figure \ref{fig:pairface} represent the skin colors and hair colors of the style target faces, while preserving the identities, poses and expressions of the content target faces. Similarly, the synthesis shoes in the third row in Figure \ref{fig:pairshoe} show the colors of the style target shoes and the structures and shapes of the content target shoes. Furthermore, for target and synthesis images, the lower the content/style perceptual distances (in Eq.~\eqref{contentloss} and Eq.~\eqref{styleloss}) are, the more similar the contents/styles are. As shown in the forth row in Figure \ref{fig:pairface} and \ref{fig:pairshoe}, the synthesis images have a lower content perceptual distance with content target images than style target images.
From the fifth row in Figure \ref{fig:pairface} and \ref{fig:pairshoe}, we could observe that the synthesis images have a lower style perceptual distances with style target images than content target images.
We conclude that the S$^3$-GAN could make the style synthesis have the following two advantages:
1) For the content target images, it only captures the content and abandons the style information.
2) For the style target images, it maintains the style and ignores the content information.

\subsubsection{Diversity}
Furthermore, we analyze the diversity of the S$^3$-GAN from the following two aspects:
1) we apply various style target images on a same content image to generate the synthesis images. As shown in Figure~\ref{fig:samplecontent}, the generated images maintain the similar structure with the original content target image and show different colors and textures according to different style target images.
2) we use a same style target image and different content target images to synthesize images. As shown in Figure \ref{fig:samplestyle}, the color and texture of the generated images are same to the style target image with different shapes and structures.
The above results present the diversity of the proposed S$^3$-GAN. In other words, S$^3$-GAN could capture various styles and contents in a single model.

\begin{figure}[t]
  \centering
  \includegraphics[width=\linewidth]{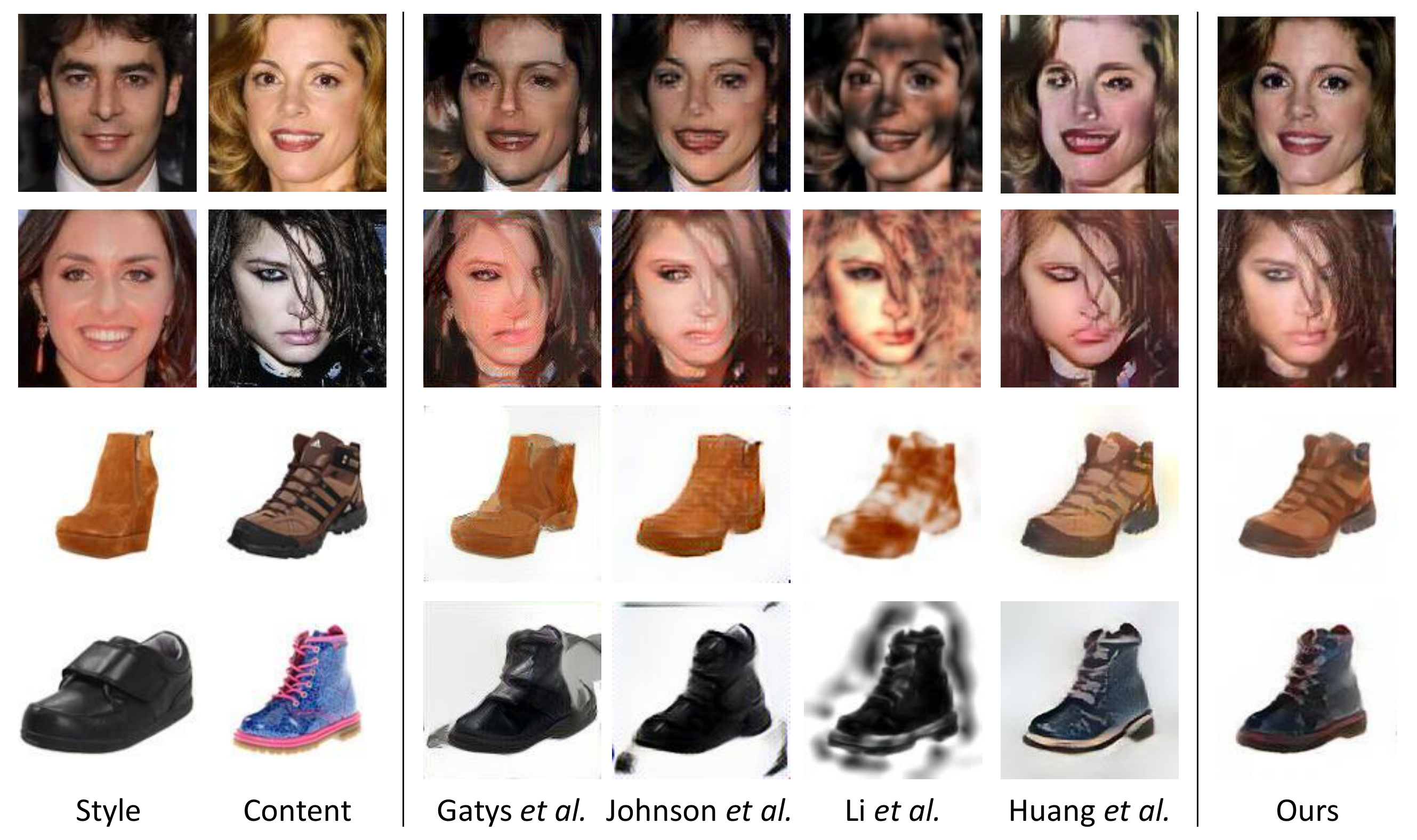}
   \vspace{-15pt}
  \caption{\small Comparison results with four popular style synthesis approaches \cite{DBLP:conf/cvpr/GatysEB16}, \cite{DBLP:conf/eccv/JohnsonAF16}, \cite{DBLP:conf/nips/LiFYWLY17} and \cite{DBLP:conf/iccv/HuangB17}.}
  \label{fig:comp}
   \vspace{-5pt}
\end{figure}

\begin{table}[t]
\caption{\small Confidence scores of S$^3$-GAN compared with four popular style synthesis approaches \cite{DBLP:conf/cvpr/GatysEB16}, \cite{DBLP:conf/eccv/JohnsonAF16}, \cite{DBLP:conf/nips/LiFYWLY17} and \cite{DBLP:conf/iccv/HuangB17}. Confidence: averaged confidence score of face detection. Content: averaged logarithms of content perceptual distances. Style: averaged logarithms of style perceptual distances.}
\vspace{-5pt}
\centering
\begin{tabular}{p{0.1cm}p{2.5cm}|p{1.7cm}<{\centering}p{1.2cm}<{\centering}p{1.2cm}<{\centering}}
\hline
& Method\centering                              &   Confidence    &   Content       &   Style \\
\hline
& Gatys \textit{et al.} \cite{DBLP:conf/cvpr/GatysEB16}   &   0.947      &    5.56    &   \textbf{3.18} \\
& Johnson \textit{et al.} \cite{DBLP:conf/eccv/JohnsonAF16} &   0.938      &    5.64    &   3.24 \\
& Li \textit{et al.} \cite{DBLP:conf/nips/LiFYWLY17}      &   0.967      &    5.66    &   3.67 \\
& Huang \textit{et al.} \cite{DBLP:conf/iccv/HuangB17}   &   0.953      &    5.36    &   3.49 \\
\hline
& S$^3$-GAN (ours)                  &   \textbf{0.984}      &    \textbf{5.25}    &   3.22 \\
\hline
\end{tabular}
   \vspace{-10pt}
\label{tab}
\end{table}

\subsubsection{Comparisons}
We compare the proposed method with other four popular style synthesis approaches \cite{DBLP:conf/cvpr/GatysEB16}, \cite{DBLP:conf/eccv/JohnsonAF16}, \cite{DBLP:conf/nips/LiFYWLY17} and \cite{DBLP:conf/iccv/HuangB17}.
As shown in Figure~\ref{fig:comp}, the images generated by S$^3$-GAN are more visually realistic because they contain distinguishable details. In contrast, the images generated by the four existing approaches are blurry and distorted, and lose many details of the content target images.
Inspired by \cite{DBLP:conf/eccv/WangG16} and \cite{DBLP:conf/nips/LiFYWLY17}, we also perform the quantitative experiment.
We randomly select 10 images as style target images and another 100 images as content target images. Then we generate 1000 synthesis images by using the five approaches.
In order to measure the realism of generated images, we employ the popular MTCNN \cite{DBLP:journals/spl/ZhangZLQ16} to perform face detection on the generated images.
The more realistic the generated image is, the higher confidence score is.
Thus, we employ the confidence score, \emph{i.e.,} the softmax probability, of face detection to represent the quality of the generated image. As shown in Table \ref{tab}, the confidence score of the proposed method is higher than the four methods, which means the generated images of the proposed method are more realistic.
Furthermore, we employ the averaged logarithms of content/style perceptual distances in Eq.~\eqref{contentloss} and Eq.~\eqref{styleloss} to measure the similarity of synthesis results and content/style target images.  As shown in Table \ref{tab}, the averaged logarithms of content perceptual distances are lower than those of the four comparison methods. Meanwhile, the averaged logarithms of style perceptual distances are lower than those of the three comparison methods \cite{DBLP:conf/eccv/JohnsonAF16}, \cite{DBLP:conf/nips/LiFYWLY17} and \cite{DBLP:conf/iccv/HuangB17}.
Note that the averaged logarithms of style perceptual distances of our method is slightly higher than \cite{DBLP:conf/cvpr/GatysEB16} because they \cite{DBLP:conf/cvpr/GatysEB16} only focus on transferring style information.
The above comparisons show that the generated images of the proposed methods could well represent the content and style of the target images.

%As shown in Figure \ref{fig:comp}, the S$^3$-GAN show better performance than them, since our generated images are more visual realistic with distinguishable details.
%By contrast, results of the four compared approaches are blurry and distorted, losing many details of the content target images.
%Moreover, the optimization-based method \cite{DBLP:conf/cvpr/GatysEB16} costs several minutes to obtain the results, which is computationally expensive. The approach based on feedforward networks \cite{DBLP:conf/eccv/JohnsonAF16} needs to learn multiple models, each for representing a style. By contrast, the proposed S$^3$-GAN only costs tens of milliseconds to generate the results through one forward propagation process, and can capture various styles in a single model.

\begin{figure}[t]
  \centering
  \includegraphics[width=0.95\linewidth]{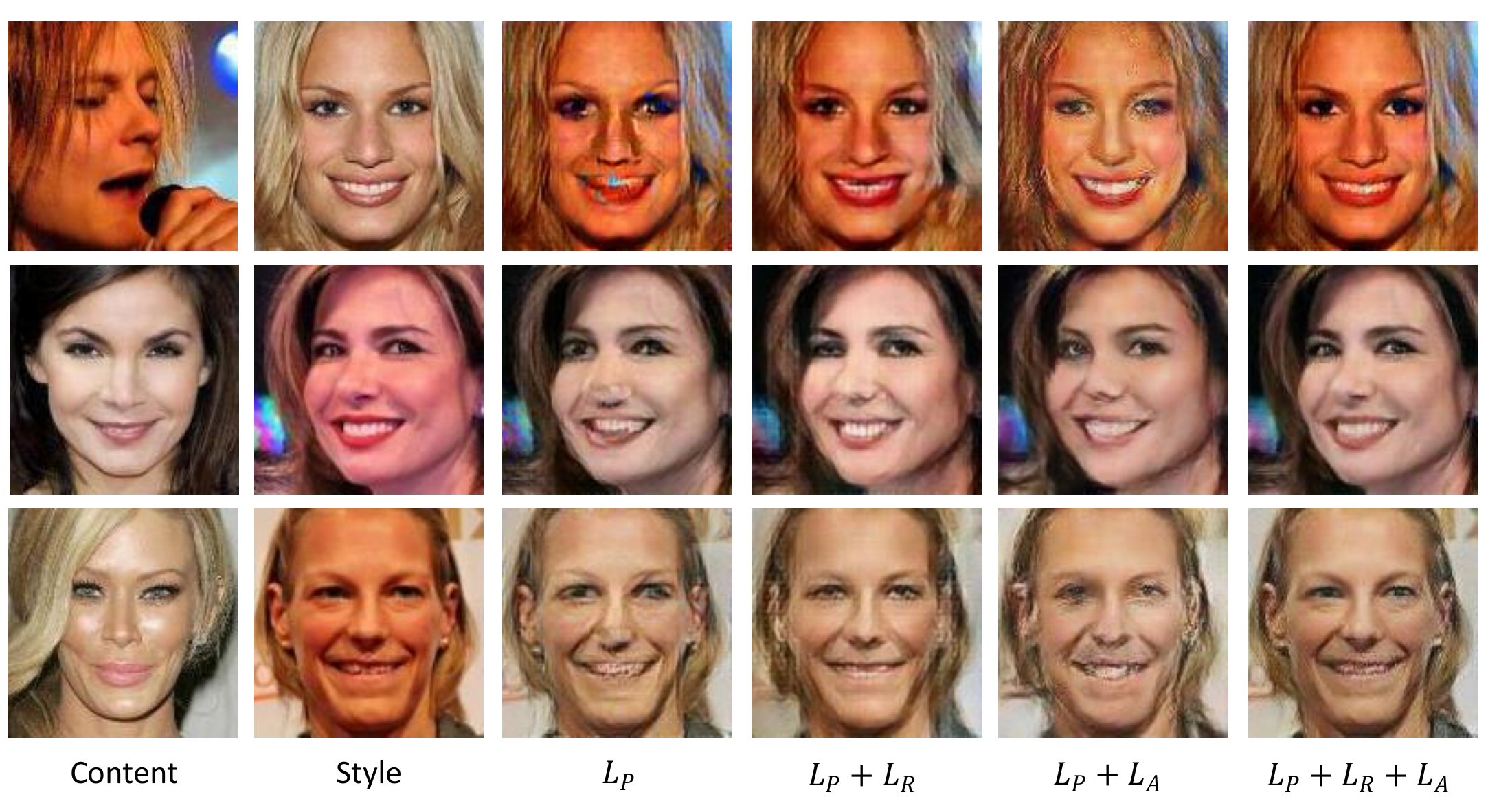}
   \vspace{-12pt}
  \caption{\small Effect of different objective functions. $L_P=L_C+L_S$ is the sum of content perceptual loss in Eq.~\eqref{contentloss} and style perceptual loss in Eq.~\eqref{styleloss}. $L_R$ is the reconstruction loss in Eq.~\eqref{reconloss}. $L_A$ is the adversary loss in Eq.~\eqref{equ8}.}
  \label{fig:obj}
   \vspace{-15pt}
\end{figure}

\subsection{Discussion and Analysis}

\subsubsection{Analysis of the objective function}

In this work, the objective function (Eq.~\eqref{fullobj}) consists of several loss functions, such as  content perceptual loss $L_{C}$, style perceptual loss $L_{S}$, reconstruction loss $L_{R}$, and adversary loss $L_{A}$. Therefore, we analyze the effect of each loss function for images, and the related results are summarized in Figure \ref{fig:obj}. From Figure \ref{fig:obj}, we could observe that the images generated by using perceptual loss $L_{P}$ ($L_P=L_C+L_S$) are blurry and distorted, and lack of many important details. The reason is that the perceptual loss is a global constraint, and has a limited ability to capture the subtle information. By adding the reconstruction loss $L_{R}$ to perceptual loss $L_{P}$, the generated images are still blurry but are more reasonable. Otherwise, the images generated by combining the adversary loss $L_{A}$ and perceptual loss $L_{P}$ are sharper and realistic, but lose much important detail information compared with the content target images. For example, the slight differences in eyes, eyebrows, and mouths between the generated and content target images could change the identities of original faces. As the reconstruction loss $L_{R}$ could ensure the encoder and generator are a pair of inverse mappings, it could compensate for the incorrect details cased by adversary loss $L_{A}$. Therefore, the reconstruction loss is complementary to adversary loss.  As shown in Figure \ref{fig:obj}, fusing the reconstruction loss $L_{R}$, adversary loss $_{A}$, and perceptual loss $L_{P}$ could tackle the above-mentioned problem, and the generated images could have sharper and corrected details.

%We analysis the importance of components in objective function in Equation ~\eqref{fullobj} through isolating the effect of adversary loss and reconstruction loss. As shown in Figure \ref{fig:obj}, results relying only on perceptual losses are blurry and distorted, lacking of important details. This is because perceptual losses are measured in a global view, with limited ability of capturing detailed information. Applying reconstruction loss along with perceptual losses will provide reasonable results, but the edges and textures are still blurry. Meanwhile, adding adversary loss to the objective function can produce sharper and realistic results. However, some important details of generated images may be different with the content target images. Namely, the eyes, eyebrows and mouths in the generated images may be slight different with the content target images, so that the identities of content target faces will be changed.
%In contrast, employing reconstruction loss and adversary loss simultaneously can tanckle the above-mentioned problems, and the generated images have sharper and corrected details. The incorrect details brought by adversary loss can be avoided with adding reconstruction loss, since reconstruction loss can ensure that the encoder and the generator are a pair of inverse mappings to each other.

\subsubsection{Analysis of Content and Style Interpolation}

\begin{figure}[t]
  \centering
  \includegraphics[width=0.95\linewidth]{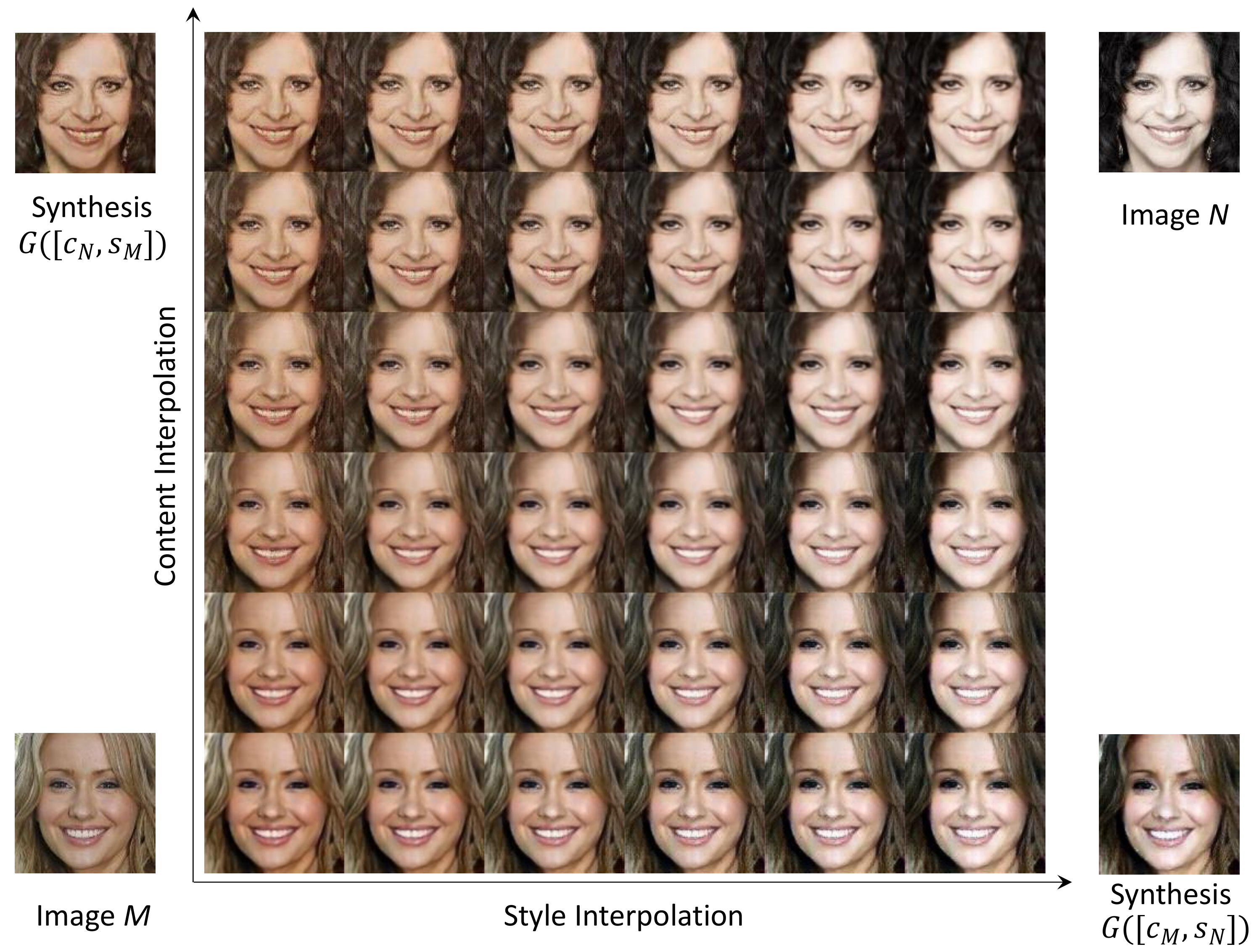}
   \vspace{-12pt}
  \caption{\small Illustration of the learned face manifold $\mathcal{M}$ and analysis of content and style interpolation. The horizontal axis indicates the traversing of style, and the vertical axis indicates the traversing of content.}
  \label{fig:interp}
   \vspace{-18pt}
\end{figure}

We analyze the assumption of manifold $\mathcal{M}$ of object photographs through illustrating the results of content and style interpolation, as shown in Figure \ref{fig:interp}. Images in the bottom left and top right corners are the reconstructions of two target faces, while images in the bottom right and top left corners are the synthesis results of swapping contents and styles. The horizontal (or vertical) axis indicates the traversing of style (or content), \emph{i.e.,} images in each row (or column) are style (or content) interpolation results with fixed content (or style). These results show that the contents and styles are independent of images in the manifold $\mathcal{M}$. Moreover, the learned encoder $E$ and generator $G$ could build mappings between the manifold $\mathcal{M}$ and the latent space $\mathcal{L}$, and successfully obtain the representation of content and style.

\section{Conclusion}

In this paper, we propose the S$^3$-GAN to implement style separation and style synthesis simultaneously. We assume the object photographs of a specific category lie on a manifold, as well as the content and style of an object are independent. We learn an encoder to build the mapping from the manifold to a latent space, in which the content and style of an object could be represented with two halves of its associated latent vector respectively. Thus, the style separation could be performed by the encoder. We also learn a generator for the inverse mapping, so that the result of style synthesis could be generated from concatenating the style half vector of the style target image and the content half vector of the content target image. Experiments on both CelebA and UT Zappos 50K datasets demonstrate the satisfactory results of the proposed S$^3$-GAN.

%% file: ms.bbl

\begin{thebibliography}{49}

%%% the Web address.

\ifx \showCODEN    \undefined \def \showCODEN     #1{\unskip}     \fi
\ifx \showDOI      \undefined \def \showDOI       #1{#1}\fi
\ifx \showISBNx    \undefined \def \showISBNx     #1{\unskip}     \fi
\ifx \showISBNxiii \undefined \def \showISBNxiii  #1{\unskip}     \fi
\ifx \showISSN     \undefined \def \showISSN      #1{\unskip}     \fi
\ifx \showLCCN     \undefined \def \showLCCN      #1{\unskip}     \fi
\ifx \shownote     \undefined \def \shownote      #1{#1}          \fi
\ifx \showarticletitle \undefined \def \showarticletitle #1{#1}   \fi
\ifx \showURL      \undefined \def \showURL       {\relax}        \fi
% invisible to TeX
\providecommand\bibfield[2]{#2}
\providecommand\bibinfo[2]{#2}
\providecommand\natexlab[1]{#1}
\providecommand\showeprint[2][]{arXiv:#2}

\bibitem[\protect\citeauthoryear{Arjovsky, Chintala, and Bottou}{Arjovsky
  et~al\mbox{.}}{2017}]%
        {DBLP:journals/corr/ArjovskyCB17}
\bibfield{author}{\bibinfo{person}{Mart{\'{\i}}n Arjovsky},
  \bibinfo{person}{Soumith Chintala}, {and} \bibinfo{person}{L{\'{e}}on
  Bottou}.} \bibinfo{year}{2017}\natexlab{}.
\newblock \showarticletitle{Wasserstein {GAN}}.
\newblock \bibinfo{journal}{\emph{arXiv preprint arXiv:1701.07875}}
  (\bibinfo{year}{2017}).
\newblock


\bibitem[\protect\citeauthoryear{Chen, Yuan, Liao, Yu, and Hua}{Chen
  et~al\mbox{.}}{2017}]%
        {DBLP:journals/corr/ChenYLYH17}
\bibfield{author}{\bibinfo{person}{Dongdong Chen}, \bibinfo{person}{Lu Yuan},
  \bibinfo{person}{Jing Liao}, \bibinfo{person}{Nenghai Yu}, {and}
  \bibinfo{person}{Gang Hua}.} \bibinfo{year}{2017}\natexlab{}.
\newblock \showarticletitle{StyleBank: An Explicit Representation for Neural
  Image Style Transfer}. In \bibinfo{booktitle}{\emph{IEEE Conference on
  Computer Vision and Pattern Recognition}}.
\newblock


\bibitem[\protect\citeauthoryear{Chi and Peng}{Chi and Peng}{2018}]%
        {DBLP:conf/ijcai/ChiP18}
\bibfield{author}{\bibinfo{person}{Jingze Chi} {and} \bibinfo{person}{Yuxin
  Peng}.} \bibinfo{year}{2018}\natexlab{}.
\newblock \showarticletitle{Dual Adversarial Networks for Zero-shot Cross-media
  Retrieval}. In \bibinfo{booktitle}{\emph{International Joint Conference on
  Artificial Intelligence}}.
\newblock


\bibitem[\protect\citeauthoryear{Deng, Dong, Socher, Li, Li, and Li}{Deng
  et~al\mbox{.}}{2009}]%
        {DBLP:conf/cvpr/DengDSLL009}
\bibfield{author}{\bibinfo{person}{Jia Deng}, \bibinfo{person}{Wei Dong},
  \bibinfo{person}{Richard Socher}, \bibinfo{person}{Li{-}Jia Li},
  \bibinfo{person}{Kai Li}, {and} \bibinfo{person}{Fei{-}Fei Li}.}
  \bibinfo{year}{2009}\natexlab{}.
\newblock \showarticletitle{ImageNet: {A} large-scale hierarchical image
  database}. In \bibinfo{booktitle}{\emph{IEEE Conference on Computer Vision
  and Pattern Recognition}}.
\newblock


\bibitem[\protect\citeauthoryear{Dumoulin, Shlens, and Kudlur}{Dumoulin
  et~al\mbox{.}}{2016}]%
        {DBLP:journals/corr/DumoulinSK16}
\bibfield{author}{\bibinfo{person}{Vincent Dumoulin}, \bibinfo{person}{Jonathon
  Shlens}, {and} \bibinfo{person}{Manjunath Kudlur}.}
  \bibinfo{year}{2016}\natexlab{}.
\newblock \showarticletitle{A Learned Representation For Artistic Style}. In
  \bibinfo{booktitle}{\emph{International Conference on Learning
  Representations}}.
\newblock


\bibitem[\protect\citeauthoryear{Efros and Freeman}{Efros and Freeman}{2001}]%
        {DBLP:conf/siggraph/EfrosF01}
\bibfield{author}{\bibinfo{person}{Alexei~A. Efros} {and}
  \bibinfo{person}{William~T. Freeman}.} \bibinfo{year}{2001}\natexlab{}.
\newblock \showarticletitle{Image quilting for texture synthesis and transfer}.
  In \bibinfo{booktitle}{\emph{Conference on Computer Graphics and Interactive
  Techniques, {SIGGRAPH}}}.
\newblock


\bibitem[\protect\citeauthoryear{Efros and Leung}{Efros and Leung}{1999}]%
        {DBLP:conf/iccv/EfrosL99}
\bibfield{author}{\bibinfo{person}{Alexei~A. Efros} {and}
  \bibinfo{person}{Thomas~K. Leung}.} \bibinfo{year}{1999}\natexlab{}.
\newblock \showarticletitle{Texture Synthesis by Non-parametric Sampling}. In
  \bibinfo{booktitle}{\emph{IEEE International Conference on Computer Vision}}.
\newblock


\bibitem[\protect\citeauthoryear{Gatys, Ecker, and Bethge}{Gatys
  et~al\mbox{.}}{2015}]%
        {DBLP:conf/nips/GatysEB15}
\bibfield{author}{\bibinfo{person}{Leon~A. Gatys},
  \bibinfo{person}{Alexander~S. Ecker}, {and} \bibinfo{person}{Matthias
  Bethge}.} \bibinfo{year}{2015}\natexlab{}.
\newblock \showarticletitle{Texture Synthesis Using Convolutional Neural
  Networks}. In \bibinfo{booktitle}{\emph{Advances in Neural Information
  Processing Systems}}.
\newblock


\bibitem[\protect\citeauthoryear{Gatys, Ecker, and Bethge}{Gatys
  et~al\mbox{.}}{2016}]%
        {DBLP:conf/cvpr/GatysEB16}
\bibfield{author}{\bibinfo{person}{Leon~A. Gatys},
  \bibinfo{person}{Alexander~S. Ecker}, {and} \bibinfo{person}{Matthias
  Bethge}.} \bibinfo{year}{2016}\natexlab{}.
\newblock \showarticletitle{Image Style Transfer Using Convolutional Neural
  Networks}. In \bibinfo{booktitle}{\emph{IEEE Conference on Computer Vision
  and Pattern Recognition}}.
\newblock


\bibitem[\protect\citeauthoryear{Ghiasi, Lee, Kudlur, Dumoulin, and
  Shlens}{Ghiasi et~al\mbox{.}}{2017}]%
        {Ghiasi2017Exploring}
\bibfield{author}{\bibinfo{person}{Golnaz Ghiasi}, \bibinfo{person}{Honglak
  Lee}, \bibinfo{person}{Manjunath Kudlur}, \bibinfo{person}{Vincent Dumoulin},
  {and} \bibinfo{person}{Jonathon Shlens}.} \bibinfo{year}{2017}\natexlab{}.
\newblock \showarticletitle{Exploring the structure of a real-time, arbitrary
  neural artistic stylization network}. In \bibinfo{booktitle}{\emph{British
  Machine Vision Conference}}.
\newblock


\bibitem[\protect\citeauthoryear{Glorot and Bengio}{Glorot and Bengio}{2010}]%
        {DBLP:journals/jmlr/GlorotB10}
\bibfield{author}{\bibinfo{person}{Xavier Glorot} {and} \bibinfo{person}{Yoshua
  Bengio}.} \bibinfo{year}{2010}\natexlab{}.
\newblock \showarticletitle{Understanding the difficulty of training deep
  feedforward neural networks}. In \bibinfo{booktitle}{\emph{International
  Conference on Artificial Intelligence and Statistics (AISTATS)}}.
\newblock


\bibitem[\protect\citeauthoryear{Goodfellow, Pouget{-}Abadie, Mirza, Xu,
  Warde{-}Farley, Ozair, Courville, and Bengio}{Goodfellow
  et~al\mbox{.}}{2014}]%
        {DBLP:conf/nips/GoodfellowPMXWOCB14}
\bibfield{author}{\bibinfo{person}{Ian~J. Goodfellow}, \bibinfo{person}{Jean
  Pouget{-}Abadie}, \bibinfo{person}{Mehdi Mirza}, \bibinfo{person}{Bing Xu},
  \bibinfo{person}{David Warde{-}Farley}, \bibinfo{person}{Sherjil Ozair},
  \bibinfo{person}{Aaron~C. Courville}, {and} \bibinfo{person}{Yoshua Bengio}.}
  \bibinfo{year}{2014}\natexlab{}.
\newblock \showarticletitle{Generative Adversarial Nets}. In
  \bibinfo{booktitle}{\emph{Advances in Neural Information Processing
  Systems}}.
\newblock


\bibitem[\protect\citeauthoryear{Gulrajani, Ahmed, Arjovsky, Dumoulin, and
  Courville}{Gulrajani et~al\mbox{.}}{2017}]%
        {DBLP:journals/corr/GulrajaniAADC17}
\bibfield{author}{\bibinfo{person}{Ishaan Gulrajani}, \bibinfo{person}{Faruk
  Ahmed}, \bibinfo{person}{Mart{\'{\i}}n Arjovsky}, \bibinfo{person}{Vincent
  Dumoulin}, {and} \bibinfo{person}{Aaron~C. Courville}.}
  \bibinfo{year}{2017}\natexlab{}.
\newblock \showarticletitle{Improved Training of Wasserstein GANs}. In
  \bibinfo{booktitle}{\emph{Advances in Neural Information Processing
  Systems}}.
\newblock


\bibitem[\protect\citeauthoryear{Hertzmann, Jacobs, Oliver, Curless, and
  Salesin}{Hertzmann et~al\mbox{.}}{2001}]%
        {DBLP:conf/siggraph/HertzmannJOCS01}
\bibfield{author}{\bibinfo{person}{Aaron Hertzmann},
  \bibinfo{person}{Charles~E. Jacobs}, \bibinfo{person}{Nuria Oliver},
  \bibinfo{person}{Brian Curless}, {and} \bibinfo{person}{David Salesin}.}
  \bibinfo{year}{2001}\natexlab{}.
\newblock \showarticletitle{Image analogies}. In
  \bibinfo{booktitle}{\emph{Conference on Computer Graphics and Interactive
  Techniques, {SIGGRAPH}}}.
\newblock


\bibitem[\protect\citeauthoryear{Huang and Belongie}{Huang and
  Belongie}{2017}]%
        {DBLP:conf/iccv/HuangB17}
\bibfield{author}{\bibinfo{person}{Xun Huang} {and} \bibinfo{person}{Serge~J.
  Belongie}.} \bibinfo{year}{2017}\natexlab{}.
\newblock \showarticletitle{Arbitrary Style Transfer in Real-Time with Adaptive
  Instance Normalization}. In \bibinfo{booktitle}{\emph{IEEE International
  Conference on Computer Vision}}.
\newblock


\bibitem[\protect\citeauthoryear{Ioffe and Szegedy}{Ioffe and Szegedy}{2015}]%
        {DBLP:conf/icml/IoffeS15}
\bibfield{author}{\bibinfo{person}{Sergey Ioffe} {and}
  \bibinfo{person}{Christian Szegedy}.} \bibinfo{year}{2015}\natexlab{}.
\newblock \showarticletitle{Batch Normalization: Accelerating Deep Network
  Training by Reducing Internal Covariate Shift}. In
  \bibinfo{booktitle}{\emph{Internal Conference on Meachine Learning}}.
\newblock


\bibitem[\protect\citeauthoryear{Isola, Zhu, Zhou, and Efros}{Isola
  et~al\mbox{.}}{2017}]%
        {DBLP:journals/corr/IsolaZZE16}
\bibfield{author}{\bibinfo{person}{Phillip Isola}, \bibinfo{person}{Jun{-}Yan
  Zhu}, \bibinfo{person}{Tinghui Zhou}, {and} \bibinfo{person}{Alexei~A.
  Efros}.} \bibinfo{year}{2017}\natexlab{}.
\newblock \showarticletitle{Image-to-Image Translation with Conditional
  Adversarial Networks}. In \bibinfo{booktitle}{\emph{IEEE Conference on
  Computer Vision and Pattern Recognition}}.
\newblock


\bibitem[\protect\citeauthoryear{Johnson, Alahi, and Fei{-}Fei}{Johnson
  et~al\mbox{.}}{2016}]%
        {DBLP:conf/eccv/JohnsonAF16}
\bibfield{author}{\bibinfo{person}{Justin Johnson}, \bibinfo{person}{Alexandre
  Alahi}, {and} \bibinfo{person}{Li Fei{-}Fei}.}
  \bibinfo{year}{2016}\natexlab{}.
\newblock \showarticletitle{Perceptual Losses for Real-Time Style Transfer and
  Super-Resolution}. In \bibinfo{booktitle}{\emph{European Conference on
  Computer Vision}}.
\newblock


\bibitem[\protect\citeauthoryear{Kim, Cha, Kim, Lee, and Kim}{Kim
  et~al\mbox{.}}{2017}]%
        {DBLP:conf/icml/KimCKLK17}
\bibfield{author}{\bibinfo{person}{Taeksoo Kim}, \bibinfo{person}{Moonsu Cha},
  \bibinfo{person}{Hyunsoo Kim}, \bibinfo{person}{Jung~Kwon Lee}, {and}
  \bibinfo{person}{Jiwon Kim}.} \bibinfo{year}{2017}\natexlab{}.
\newblock \showarticletitle{Learning to Discover Cross-Domain Relations with
  Generative Adversarial Networks}. In \bibinfo{booktitle}{\emph{Internal
  Conference on Meachine Learning}}.
\newblock


\bibitem[\protect\citeauthoryear{Kingma and Ba}{Kingma and Ba}{2015}]%
        {DBLP:journals/corr/KingmaB14}
\bibfield{author}{\bibinfo{person}{Diederik~P. Kingma} {and}
  \bibinfo{person}{Jimmy Ba}.} \bibinfo{year}{2015}\natexlab{}.
\newblock \showarticletitle{Adam: {A} Method for Stochastic Optimization}.
\newblock \bibinfo{journal}{\emph{International Conference on Learning
  Representations}}.
\newblock


\bibitem[\protect\citeauthoryear{Krizhevsky, Sutskever, and Hinton}{Krizhevsky
  et~al\mbox{.}}{2012}]%
        {DBLP:conf/nips/KrizhevskySH12}
\bibfield{author}{\bibinfo{person}{Alex Krizhevsky}, \bibinfo{person}{Ilya
  Sutskever}, {and} \bibinfo{person}{Geoffrey~E. Hinton}.}
  \bibinfo{year}{2012}\natexlab{}.
\newblock \showarticletitle{ImageNet Classification with Deep Convolutional
  Neural Networks}. In \bibinfo{booktitle}{\emph{Advances in Neural Information
  Processing Systems}}.
\newblock


\bibitem[\protect\citeauthoryear{Ledig, Theis, Huszar, Caballero, Aitken,
  Tejani, Totz, Wang, and Shi}{Ledig et~al\mbox{.}}{2017}]%
        {DBLP:journals/corr/LedigTHCATTWS16}
\bibfield{author}{\bibinfo{person}{Christian Ledig}, \bibinfo{person}{Lucas
  Theis}, \bibinfo{person}{Ferenc Huszar}, \bibinfo{person}{Jose Caballero},
  \bibinfo{person}{Andrew~P. Aitken}, \bibinfo{person}{Alykhan Tejani},
  \bibinfo{person}{Johannes Totz}, \bibinfo{person}{Zehan Wang}, {and}
  \bibinfo{person}{Wenzhe Shi}.} \bibinfo{year}{2017}\natexlab{}.
\newblock \showarticletitle{Photo-Realistic Single Image Super-Resolution Using
  a Generative Adversarial Network}. In \bibinfo{booktitle}{\emph{IEEE
  Conference on Computer Vision and Pattern Recognition}}.
\newblock


\bibitem[\protect\citeauthoryear{Li and Wand}{Li and Wand}{2016a}]%
        {DBLP:conf/cvpr/LiW16}
\bibfield{author}{\bibinfo{person}{Chuan Li} {and} \bibinfo{person}{Michael
  Wand}.} \bibinfo{year}{2016}\natexlab{a}.
\newblock \showarticletitle{Combining Markov Random Fields and Convolutional
  Neural Networks for Image Synthesis}. In \bibinfo{booktitle}{\emph{IEEE
  Conference on Computer Vision and Pattern Recognition}}.
\newblock


\bibitem[\protect\citeauthoryear{Li and Wand}{Li and Wand}{2016b}]%
        {DBLP:conf/eccv/LiW16}
\bibfield{author}{\bibinfo{person}{Chuan Li} {and} \bibinfo{person}{Michael
  Wand}.} \bibinfo{year}{2016}\natexlab{b}.
\newblock \showarticletitle{Precomputed Real-Time Texture Synthesis with
  Markovian Generative Adversarial Networks}. In
  \bibinfo{booktitle}{\emph{European Conference on Computer Vision}}.
\newblock


\bibitem[\protect\citeauthoryear{Li, Fang, Yang, Wang, Lu, and Yang}{Li
  et~al\mbox{.}}{2017a}]%
        {DBLP:conf/cvpr/LiFYWLY17}
\bibfield{author}{\bibinfo{person}{Yijun Li}, \bibinfo{person}{Chen Fang},
  \bibinfo{person}{Jimei Yang}, \bibinfo{person}{Zhaowen Wang},
  \bibinfo{person}{Xin Lu}, {and} \bibinfo{person}{Ming{-}Hsuan Yang}.}
  \bibinfo{year}{2017}\natexlab{a}.
\newblock \showarticletitle{Diversified Texture Synthesis with Feed-Forward
  Networks}. In \bibinfo{booktitle}{\emph{IEEE Conference on Computer Vision
  and Pattern Recognition}}.
\newblock


\bibitem[\protect\citeauthoryear{Li, Fang, Yang, Wang, Lu, and Yang}{Li
  et~al\mbox{.}}{2017b}]%
        {DBLP:conf/nips/LiFYWLY17}
\bibfield{author}{\bibinfo{person}{Yijun Li}, \bibinfo{person}{Chen Fang},
  \bibinfo{person}{Jimei Yang}, \bibinfo{person}{Zhaowen Wang},
  \bibinfo{person}{Xin Lu}, {and} \bibinfo{person}{Ming{-}Hsuan Yang}.}
  \bibinfo{year}{2017}\natexlab{b}.
\newblock \showarticletitle{Universal Style Transfer via Feature Transforms}.
  In \bibinfo{booktitle}{\emph{Advances in Neural Information Processing
  Systems}}.
\newblock


\bibitem[\protect\citeauthoryear{Liu, Luo, Wang, and Tang}{Liu
  et~al\mbox{.}}{2015}]%
        {DBLP:conf/iccv/LiuLWT15}
\bibfield{author}{\bibinfo{person}{Ziwei Liu}, \bibinfo{person}{Ping Luo},
  \bibinfo{person}{Xiaogang Wang}, {and} \bibinfo{person}{Xiaoou Tang}.}
  \bibinfo{year}{2015}\natexlab{}.
\newblock \showarticletitle{Deep Learning Face Attributes in the Wild}. In
  \bibinfo{booktitle}{\emph{IEEE International Conference on Computer Vision}}.
\newblock


\bibitem[\protect\citeauthoryear{Ma, Jia, Sun, Schiele, Tuytelaars, and
  Gool}{Ma et~al\mbox{.}}{2017}]%
        {DBLP:conf/nips/MaJSSTG17}
\bibfield{author}{\bibinfo{person}{Liqian Ma}, \bibinfo{person}{Xu Jia},
  \bibinfo{person}{Qianru Sun}, \bibinfo{person}{Bernt Schiele},
  \bibinfo{person}{Tinne Tuytelaars}, {and} \bibinfo{person}{Luc~Van Gool}.}
  \bibinfo{year}{2017}\natexlab{}.
\newblock \showarticletitle{Pose Guided Person Image Generation}. In
  \bibinfo{booktitle}{\emph{Advances in Neural Information Processing
  Systems}}.
\newblock


\bibitem[\protect\citeauthoryear{Mirza and Osindero}{Mirza and
  Osindero}{2014}]%
        {DBLP:journals/corr/MirzaO14}
\bibfield{author}{\bibinfo{person}{Mehdi Mirza} {and} \bibinfo{person}{Simon
  Osindero}.} \bibinfo{year}{2014}\natexlab{}.
\newblock \showarticletitle{Conditional Generative Adversarial Nets}.
\newblock \bibinfo{journal}{\emph{arXiv preprint arXiv:1411.1784}}
  (\bibinfo{year}{2014}).
\newblock


\bibitem[\protect\citeauthoryear{Pathak, Kr{\"{a}}henb{\"{u}}hl, Donahue,
  Darrell, and Efros}{Pathak et~al\mbox{.}}{2016}]%
        {DBLP:conf/cvpr/PathakKDDE16}
\bibfield{author}{\bibinfo{person}{Deepak Pathak}, \bibinfo{person}{Philipp
  Kr{\"{a}}henb{\"{u}}hl}, \bibinfo{person}{Jeff Donahue},
  \bibinfo{person}{Trevor Darrell}, {and} \bibinfo{person}{Alexei~A. Efros}.}
  \bibinfo{year}{2016}\natexlab{}.
\newblock \showarticletitle{Context Encoders: Feature Learning by Inpainting}.
  In \bibinfo{booktitle}{\emph{IEEE Conference on Computer Vision and Pattern
  Recognition}}.
\newblock


\bibitem[\protect\citeauthoryear{Qi}{Qi}{2017}]%
        {DBLP:journals/corr/Qi17}
\bibfield{author}{\bibinfo{person}{Guo{-}Jun Qi}.}
  \bibinfo{year}{2017}\natexlab{}.
\newblock \showarticletitle{Loss-Sensitive Generative Adversarial Networks on
  Lipschitz Densities}.
\newblock \bibinfo{journal}{\emph{arXiv preprint arXiv:1701.06264}}
  (\bibinfo{year}{2017}).
\newblock


\bibitem[\protect\citeauthoryear{Radford, Metz, and Chintala}{Radford
  et~al\mbox{.}}{2016}]%
        {DBLP:journals/corr/RadfordMC15}
\bibfield{author}{\bibinfo{person}{Alec Radford}, \bibinfo{person}{Luke Metz},
  {and} \bibinfo{person}{Soumith Chintala}.} \bibinfo{year}{2016}\natexlab{}.
\newblock \showarticletitle{Unsupervised Representation Learning with Deep
  Convolutional Generative Adversarial Networks}. In
  \bibinfo{booktitle}{\emph{International Conference on Learning
  Representations}}.
\newblock


\bibitem[\protect\citeauthoryear{Reed, Akata, Yan, Logeswaran, Schiele, and
  Lee}{Reed et~al\mbox{.}}{2016}]%
        {DBLP:conf/icml/ReedAYLSL16}
\bibfield{author}{\bibinfo{person}{Scott~E. Reed}, \bibinfo{person}{Zeynep
  Akata}, \bibinfo{person}{Xinchen Yan}, \bibinfo{person}{Lajanugen
  Logeswaran}, \bibinfo{person}{Bernt Schiele}, {and} \bibinfo{person}{Honglak
  Lee}.} \bibinfo{year}{2016}\natexlab{}.
\newblock \showarticletitle{Generative Adversarial Text to Image Synthesis}. In
  \bibinfo{booktitle}{\emph{Internal Conference on Meachine Learning}}.
\newblock


\bibitem[\protect\citeauthoryear{Shen and Liu}{Shen and Liu}{2017}]%
        {DBLP:journals/corr/ShenL16b}
\bibfield{author}{\bibinfo{person}{Wei Shen} {and} \bibinfo{person}{Rujie
  Liu}.} \bibinfo{year}{2017}\natexlab{}.
\newblock \showarticletitle{Learning Residual Images for Face Attribute
  Manipulation}. In \bibinfo{booktitle}{\emph{IEEE Conference on Computer
  Vision and Pattern Recognition}}.
\newblock


\bibitem[\protect\citeauthoryear{Siarohin, Sangineto, Lathuili{\`{e}}re, and
  Sebe}{Siarohin et~al\mbox{.}}{2018}]%
        {DBLP:journals/corr/abs-1801-00055}
\bibfield{author}{\bibinfo{person}{Aliaksandr Siarohin}, \bibinfo{person}{Enver
  Sangineto}, \bibinfo{person}{St{\'{e}}phane Lathuili{\`{e}}re}, {and}
  \bibinfo{person}{Nicu Sebe}.} \bibinfo{year}{2018}\natexlab{}.
\newblock \showarticletitle{Deformable GANs for Pose-based Human Image
  Generation}.
\newblock \bibinfo{journal}{\emph{arXiv preprint arXiv:1801.00055}}
  (\bibinfo{year}{2018}).
\newblock


\bibitem[\protect\citeauthoryear{Simonyan and Zisserman}{Simonyan and
  Zisserman}{2014}]%
        {DBLP:journals/corr/SimonyanZ14a}
\bibfield{author}{\bibinfo{person}{Karen Simonyan} {and}
  \bibinfo{person}{Andrew Zisserman}.} \bibinfo{year}{2014}\natexlab{}.
\newblock \showarticletitle{Very Deep Convolutional Networks for Large-Scale
  Image Recognition}.
\newblock \bibinfo{journal}{\emph{arXiv preprint arXiv:1409.1556}}
  (\bibinfo{year}{2014}).
\newblock


\bibitem[\protect\citeauthoryear{Tsai, Hung, Schulter, Sohn, Yang, and
  Chandraker}{Tsai et~al\mbox{.}}{2018}]%
        {DBLP:journals/corr/abs-1802-10349}
\bibfield{author}{\bibinfo{person}{Yi{-}Hsuan Tsai},
  \bibinfo{person}{Wei{-}Chih Hung}, \bibinfo{person}{Samuel Schulter},
  \bibinfo{person}{Kihyuk Sohn}, \bibinfo{person}{Ming{-}Hsuan Yang}, {and}
  \bibinfo{person}{Manmohan Chandraker}.} \bibinfo{year}{2018}\natexlab{}.
\newblock \showarticletitle{Learning to Adapt Structured Output Space for
  Semantic Segmentation}. In \bibinfo{booktitle}{\emph{IEEE Conference on
  Computer Vision and Pattern Recognition}}.
\newblock


\bibitem[\protect\citeauthoryear{Ulyanov, Lebedev, Vedaldi, and
  Lempitsky}{Ulyanov et~al\mbox{.}}{2016}]%
        {DBLP:conf/icml/UlyanovLVL16}
\bibfield{author}{\bibinfo{person}{Dmitry Ulyanov}, \bibinfo{person}{Vadim
  Lebedev}, \bibinfo{person}{Andrea Vedaldi}, {and} \bibinfo{person}{Victor~S.
  Lempitsky}.} \bibinfo{year}{2016}\natexlab{}.
\newblock \showarticletitle{Texture Networks: Feed-forward Synthesis of
  Textures and Stylized Images}. In \bibinfo{booktitle}{\emph{Internal
  Conference on Meachine Learning}}.
\newblock


\bibitem[\protect\citeauthoryear{Wang and Gupta}{Wang and Gupta}{2016}]%
        {DBLP:conf/eccv/WangG16}
\bibfield{author}{\bibinfo{person}{Xiaolong Wang} {and}
  \bibinfo{person}{Abhinav Gupta}.} \bibinfo{year}{2016}\natexlab{}.
\newblock \showarticletitle{Generative Image Modeling Using Style and Structure
  Adversarial Networks}. In \bibinfo{booktitle}{\emph{European Conference on
  Computer Vision}}.
\newblock


\bibitem[\protect\citeauthoryear{Yang, Lu, Lin, Shechtman, Wang, and Li}{Yang
  et~al\mbox{.}}{2017}]%
        {DBLP:journals/corr/YangLLSWL16}
\bibfield{author}{\bibinfo{person}{Chao Yang}, \bibinfo{person}{Xin Lu},
  \bibinfo{person}{Zhe Lin}, \bibinfo{person}{Eli Shechtman},
  \bibinfo{person}{Oliver Wang}, {and} \bibinfo{person}{Hao Li}.}
  \bibinfo{year}{2017}\natexlab{}.
\newblock \showarticletitle{High-Resolution Image Inpainting using Multi-Scale
  Neural Patch Synthesis}. In \bibinfo{booktitle}{\emph{IEEE Conference on
  Computer Vision and Pattern Recognition}}.
\newblock


\bibitem[\protect\citeauthoryear{Yao, Zhang, Zhang, Li, and Tian}{Yao
  et~al\mbox{.}}{2017}]%
        {DBLP:conf/mm/YaoZZLT17}
\bibfield{author}{\bibinfo{person}{Hantao Yao}, \bibinfo{person}{Shiliang
  Zhang}, \bibinfo{person}{Yongdong Zhang}, \bibinfo{person}{Jintao Li}, {and}
  \bibinfo{person}{Qi Tian}.} \bibinfo{year}{2017}\natexlab{}.
\newblock \showarticletitle{One-Shot Fine-Grained Instance Retrieval}. In
  \bibinfo{booktitle}{\emph{ACM Multimedia Conference}}.
\newblock


\bibitem[\protect\citeauthoryear{Yu and Grauman}{Yu and Grauman}{2014}]%
        {DBLP:conf/cvpr/YuG14}
\bibfield{author}{\bibinfo{person}{Aron Yu} {and} \bibinfo{person}{Kristen
  Grauman}.} \bibinfo{year}{2014}\natexlab{}.
\newblock \showarticletitle{Fine-Grained Visual Comparisons with Local
  Learning}. In \bibinfo{booktitle}{\emph{IEEE Conference on Computer Vision
  and Pattern Recognition}}.
\newblock


\bibitem[\protect\citeauthoryear{Zhang, Peng, and Yuan}{Zhang
  et~al\mbox{.}}{2018a}]%
        {DBLP:conf/aaai/ZhangPY18}
\bibfield{author}{\bibinfo{person}{Jian Zhang}, \bibinfo{person}{Yuxin Peng},
  {and} \bibinfo{person}{Mingkuan Yuan}.} \bibinfo{year}{2018}\natexlab{a}.
\newblock \showarticletitle{Unsupervised Generative Adversarial Cross-Modal
  Hashing}. In \bibinfo{booktitle}{\emph{The Thirty-Second {AAAI} Conference on
  Artificial Intelligence}}.
\newblock


\bibitem[\protect\citeauthoryear{Zhang, Zhang, Li, and Qiao}{Zhang
  et~al\mbox{.}}{2016}]%
        {DBLP:journals/spl/ZhangZLQ16}
\bibfield{author}{\bibinfo{person}{Kaipeng Zhang}, \bibinfo{person}{Zhanpeng
  Zhang}, \bibinfo{person}{Zhifeng Li}, {and} \bibinfo{person}{Yu Qiao}.}
  \bibinfo{year}{2016}\natexlab{}.
\newblock \showarticletitle{Joint Face Detection and Alignment Using Multitask
  Cascaded Convolutional Networks}.
\newblock \bibinfo{journal}{\emph{{IEEE} Signal Process. Lett.}}
  (\bibinfo{year}{2016}).
\newblock


\bibitem[\protect\citeauthoryear{Zhang, Qiu, Yao, Liu, and Mei}{Zhang
  et~al\mbox{.}}{2018b}]%
        {DBLP:journals/corr/abs-1804-08286}
\bibfield{author}{\bibinfo{person}{Yiheng Zhang}, \bibinfo{person}{Zhaofan
  Qiu}, \bibinfo{person}{Ting Yao}, \bibinfo{person}{Dong Liu}, {and}
  \bibinfo{person}{Tao Mei}.} \bibinfo{year}{2018}\natexlab{b}.
\newblock \showarticletitle{Fully Convolutional Adaptation Networks for
  Semantic Segmentation}.
\newblock  (\bibinfo{year}{2018}).
\newblock


\bibitem[\protect\citeauthoryear{Zhang, Song, and Qi}{Zhang
  et~al\mbox{.}}{2017}]%
        {DBLP:journals/corr/ZhangSQ17}
\bibfield{author}{\bibinfo{person}{Zhifei Zhang}, \bibinfo{person}{Yang Song},
  {and} \bibinfo{person}{Hairong Qi}.} \bibinfo{year}{2017}\natexlab{}.
\newblock \showarticletitle{Age Progression/Regression by Conditional
  Adversarial Autoencoder}. In \bibinfo{booktitle}{\emph{IEEE Conference on
  Computer Vision and Pattern Recognition}}.
\newblock


\bibitem[\protect\citeauthoryear{Zhao, Mathieu, and LeCun}{Zhao
  et~al\mbox{.}}{2016}]%
        {DBLP:journals/corr/ZhaoML16}
\bibfield{author}{\bibinfo{person}{Junbo~Jake Zhao},
  \bibinfo{person}{Micha{\"{e}}l Mathieu}, {and} \bibinfo{person}{Yann LeCun}.}
  \bibinfo{year}{2016}\natexlab{}.
\newblock \showarticletitle{Energy-based Generative Adversarial Network}.
\newblock \bibinfo{journal}{\emph{arXiv preprint arXiv:1609.03126}}
  (\bibinfo{year}{2016}).
\newblock


\bibitem[\protect\citeauthoryear{Zhou, Xiao, Yang, Feng, He, and He}{Zhou
  et~al\mbox{.}}{2017}]%
        {DBLP:journals/corr/ZhouXYFHH17}
\bibfield{author}{\bibinfo{person}{Shuchang Zhou}, \bibinfo{person}{Taihong
  Xiao}, \bibinfo{person}{Yi Yang}, \bibinfo{person}{Dieqiao Feng},
  \bibinfo{person}{Qinyao He}, {and} \bibinfo{person}{Weiran He}.}
  \bibinfo{year}{2017}\natexlab{}.
\newblock \showarticletitle{GeneGAN: Learning Object Transfiguration and
  Attribute Subspace from Unpaired Data}.
\newblock \bibinfo{journal}{\emph{arXiv preprint arXiv:1705.04932}}
  (\bibinfo{year}{2017}).
\newblock


\bibitem[\protect\citeauthoryear{Zhu, Park, Isola, and Efros}{Zhu
  et~al\mbox{.}}{2017}]%
        {DBLP:journals/corr/ZhuPIE17}
\bibfield{author}{\bibinfo{person}{Jun{-}Yan Zhu}, \bibinfo{person}{Taesung
  Park}, \bibinfo{person}{Phillip Isola}, {and} \bibinfo{person}{Alexei~A.
  Efros}.} \bibinfo{year}{2017}\natexlab{}.
\newblock \showarticletitle{Unpaired Image-to-Image Translation using
  Cycle-Consistent Adversarial Networks}. In \bibinfo{booktitle}{\emph{IEEE
  International Conference on Computer Vision}}.
\newblock


\end{thebibliography}
